%% file: arxiv.tex
\documentclass[10pt,twocolumn,letterpaper]{article}

\usepackage[pagenumbers]{cvpr} 

\usepackage[most]{tcolorbox}
\usepackage{enumitem}
\usepackage{listings}
\usepackage{xcolor}
\usepackage{pifont}
\usepackage[table]{xcolor}
\usepackage{booktabs}
\usepackage{multirow}
\usepackage{array}
\usepackage[T1]{fontenc}
\usepackage{amsthm}
\newtheorem{definition}{Definition}

\newcommand{\cmark}{\textcolor{green}{\ding{51}}} 
\newcommand{\xmark}{\textcolor{gray}{\ding{55}}}  
\newcommand{\pmark}{%
  \textcolor{orange}{%
    \ding{51}
    \kern-0.65em
    \raisebox{0.05em}{\ding{55}}
  }%
}

\lstdefinelanguage{json}{
    basicstyle=\ttfamily\small,
    showstringspaces=false,
    breaklines=true,
    frame=single,
    backgroundcolor=\color{gray!10},
    literate=
     *{0}{{{\color{numb}0}}}{1}
      {1}{{{\color{numb}1}}}{1}
      {2}{{{\color{numb}2}}}{1}
      {3}{{{\color{numb}3}}}{1}
      {4}{{{\color{numb}4}}}{1}
      {5}{{{\color{numb}5}}}{1}
      {6}{{{\color{numb}6}}}{1}
      {7}{{{\color{numb}7}}}{1}
      {8}{{{\color{numb}8}}}{1}
      {9}{{{\color{numb}9}}}{1}
}

\definecolor{numb}{rgb}{0.6,0.0,0.0}
\definecolor{darkgreen}{RGB}{0,100,0}
\input{preamble}
\definecolor{cvprblue}{rgb}{0.21,0.49,0.74}
\usepackage[pagebackref,breaklinks,colorlinks,allcolors=cvprblue]{hyperref}
\usepackage{hyperref} 
\def\paperID{6006} 
\def\confName{CVPR}
\def\confYear{2026}

\title{ZoomEarth: Active Perception for Ultra-High-Resolution Geospatial Vision-Language Tasks}

\author{
Ruixun Liu$^{1*}$, Bowen Fu$^{1*}$, Jiayi Song$^{1}$, Kaiyu Li$^{1}$, Wanchen Li$^{1}$, Lanxuan Xue$^{1}$,\\
Hui Qiao$^{2}$, Weizhan Zhang$^{1}$, Deyu Meng$^{1}$, Xiangyong Cao$^{1\dagger}$\\[4pt]
$^{1}$Xi'an Jiaotong University \quad $^{2}$ China Telecom Shaanxi Branch\\[2pt]
\textbf{Project:} \href{https://earth-insights.github.io/ZoomEarth}{\textcolor{magenta}{https://earth-insights.github.io/ZoomEarth}}
}

\begin{document}

\twocolumn[{%
\renewcommand\twocolumn[1][]{#1}
\maketitle

\vspace{-12mm}
    \begin{center}
    \centering
    \captionsetup{type=figure}
    \includegraphics[width=1.0\linewidth]{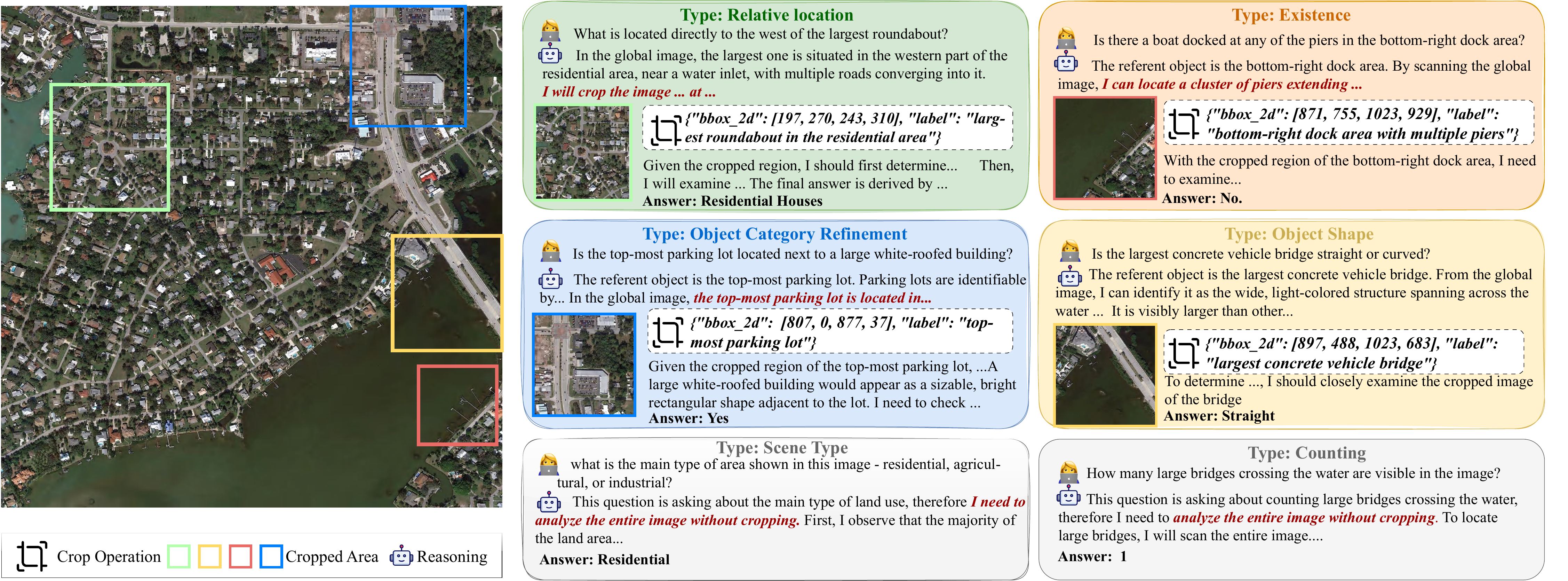}
    \vspace{-6mm}
    \captionof{figure}{Typical examples from our proposed benchmark LRS-GRO and results obtained by our ZoomEarth framework. LRS-GRO focuses on UHR RS imagery, including 17  multimodal vision-language understanding categories and emphasizing active perception process.}
    \label{fig:fig_title}
\end{center}%
 \vspace{-2mm}
}]

\renewcommand{\thefootnote}{*}
\footnotetext[1]{Equal contribution}
\renewcommand{\thefootnote}{\dag}
\footnotetext[2]{Corresponding author (caoxiangyong@mail.xjtu.edu.cn)}
\renewcommand{\thefootnote}{\arabic{footnote}}

\input{sec/0_abstract}

\input{sec/1_intro}

\input{sec/2_relatedwork}
\input{sec/3_method}

\input{sec/4_experiment}

\input{sec/5_conclusion}

\newpage
{
    \small
    \bibliographystyle{ieeenat_fullname}
    \bibliography{main}
}

\input{sec/X_suppl}

\end{document}

%% file: sec/0_abstract.tex
\begin{abstract}
Ultra-high-resolution (UHR) remote sensing (RS) images offer rich fine-grained information but also present challenges in effective processing. Existing dynamic resolution and token pruning methods are constrained by a passive perception paradigm, suffering from increased redundancy when obtaining finer visual inputs. In this work, we explore a new active perception paradigm that enables models to revisit information-rich regions. First, we present LRS-GRO, a large-scale benchmark dataset tailored for active perception in UHR RS processing, encompassing 17 question types across global, region, and object levels, annotated via a semi-automatic pipeline. Building on LRS-GRO, we propose ZoomEarth, an adaptive cropping–zooming framework with a novel Region-Guided reward that provides fine-grained guidance. Trained via supervised fine-tuning (SFT) and Group Relative Policy Optimization (GRPO), ZoomEarth achieves state-of-the-art performance on LRS-GRO and, in the zero-shot setting, on three public UHR remote sensing benchmarks. Furthermore, ZoomEarth can be seamlessly integrated with downstream models for tasks  such as cloud removal, denoising, segmentation, and image editing through simple tool interfaces, demonstrating strong versatility and extensibility.
\end{abstract}

%% file: sec/1_intro.tex
 \vspace{-2mm}
\section{Introduction}

\label{sec:intro}


Multimodal large language models (MLLMs)~\cite{liu2023llava,liu2024llavanext,beyer2024paligemmaversatile3bvlm,geminiteam2025geminifamilyhighlycapable} have been developing rapidly, demonstrating remarkable  capabilities in multimodal understanding and perception. Several studies~\cite{kuckreja2023geochat,mall2023remotesens,li2025segearth,Weng_2025} have further explored and applied vision-language models (VLMs) to RS images, achieving notable results in tasks such as visual question answering (VQA), grounding, and segmentation. However, satellite and aerial images often cover vast geographic areas, presenting significant challenges in efficiently inputting and processing UHR RS data.

\begin{figure}[t]
    \centering
    \includegraphics[width=0.98\linewidth]{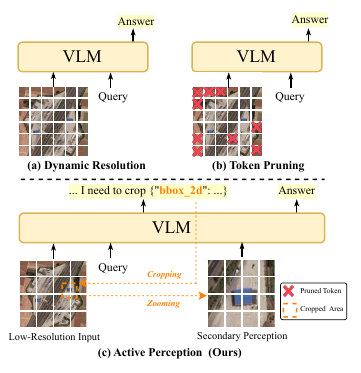}
    \vspace{-6mm}
    \caption{Comparison between passive perception and active perception. (a) Dynamic Resolution~\cite{dong2024internlmxcomposer24khdpioneeringlargevisionlanguage, zhao2024dynrefer} and (b) Token Pruning~\cite{guo2025cropcontextualregionorientedvisual, Alvar_2025_CVPR} represent passive perception approaches with only a single image input. (c) We introduce a cropping–zooming based active perception method. ``Zooming'' refers to restoring the cropped image at its original high-resolution image.}
    \label{contrast}
\vspace{-5mm}
\end{figure}

To address this, dynamic-resolution-based methods~\cite{li2024tokenpackerefficientvisualprojector,Liu_2024_CVPR,wang2024qwen2, zhou2025dynrslvlmenhancingautonomousdriving} are developed, enabling models to process high-resolution inputs directly but introduce significant computational cost. Token pruning methods~\citep{wang2025geollava8kscalingremotesensingmultimodal, luo2025largevisionlanguagemodelmeets} enhance this process via manually crafted rules (\eg, removing background tokens through clustering), yet they often fail to generalize to RS images with complex backgrounds. As shown in Fig.~\ref{contrast}, these studies remain constrained to a passive perception paradigm that relies solely on a single visual input. Specifically, when attempting to acquire more fine-grained visual representations, the model requires higher-resolution image inputs, which in turn forces it to cope with redundant visual information interference.


In this work, to overcome the limitation introduced by passive perception, we explore a new active perception paradigm that enables models to revisit regions relevant to the given query. However, implementing active perception in UHR RS faces two key challenges:
(1) the lack of RS image datasets that explicitly capture the active perception process, and
(2) the difficulty of enabling models to perform adaptive region selection and active exploration.
Recent studies~\citep{jiang2025vlmr3regionrecognitionreasoning,wang2025vgrvisualgroundedreasoning} mainly rely on optical character recognition (OCR) and object detection to construct supervised datasets, thereby achieving active perception through tool invocation. However, such approaches are not applicable to the RS domain, which typically lacks textual cues, and features targets that are often expansive regions (\ie, the industrial zone)  rather than solely discrete object instances.



To stimulate and enhance active perception in RS, we first present LRS-GRO, a large-scale benchmark dataset tailored for this paradigm. To explicitly model and supervise the active search process, the detail-oriented questions in LRS-GRO are annotated with precise bounding boxes (BBoxes) for the regions of interest (ROIs). In addition, for the supervised fine-tuning (SFT) dataset, each sample is accompanied by a Chain-of-Thought (CoT) that guides the model to first localize the ROI and then reason about the cropped details. To ensure the fidelity of these annotations, we introduce a novel semi-automatic annotation pipeline, which overcomes the defects observed in automated GPT-based workflows~\cite{luo2025largevisionlanguagemodelmeets} on UHR RS images. Leveraging this pipeline, LRS-GRO encompasses 17 question types across global, region, and object levels, providing a comprehensive benchmark that evaluates not only VQA accuracy but also the active localization of correct ROIs.


\begin{figure*}
    \centering
    \includegraphics[width=0.95\linewidth]{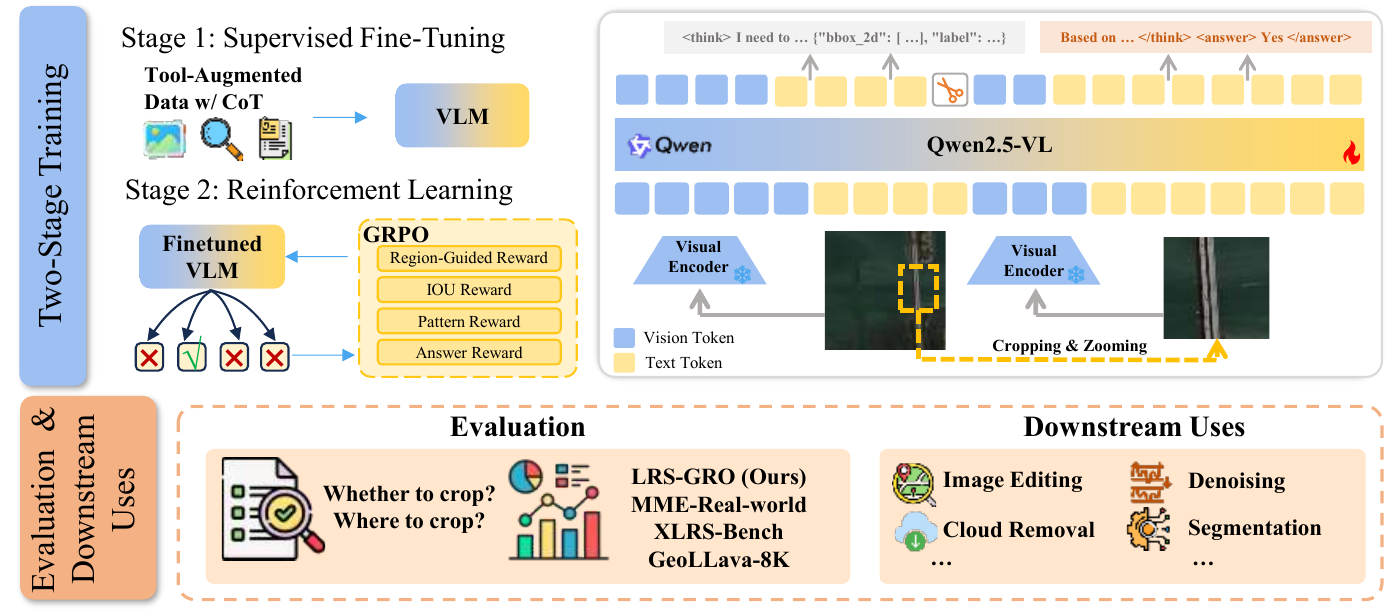}
    \vspace{-4mm}
    \caption{The visualization of the training and evaluation pipeline of our proposed methods. The model architecture diagram in the upper-right corner demonstrates the model's ability to adaptively crop the ROI by generating the BBox, and subsequently perform advanced reasoning. For clarity in the illustration, we omit the input of query tokens.}
    \label{flowchart}
    \vspace{-5mm}
\end{figure*}


Building on this foundation, we propose ZoomEarth, an adaptive cropping-zooming MLLM framework designed for active perception in UHR RS. The model first processes a downsampled global image to gain a holistic understanding. It then actively identifies ROIs, represented as BBoxes. These ROIs are subsequently cropped from the original high-resolution image and re-fed into the model for a detailed perception. As shown in Fig.~\ref{fig:fig_title}, ZoomEarth can answer global queries from the initial view while adaptively ``zooming-in'' to handle region and object-specific queries.


The framework is trained using a two-stage strategy as shown in Fig.~\ref{flowchart}. First, SFT process teaches the model the basic localization-cropping-answering thought pattern. Second, we employ GRPO~\citep{shao2024deepseekmathpushinglimitsmathematical} to enhance its reasoning capabilities. Recognizing that standard IoU-based rewards are often too sparse for UHR images, we introduce a novel Region-Guided reward. This reward leverages the spatial distribution characteristics of RS scenes, providing denser and more fine-grained guidance by rewarding predictions based on their distance to the Ground Truth (GT).

Our main contributions are summarized as follows:
\begin{itemize}
    \item We propose a UHR RS benchmark dataset, LRS-GRO, which encompasses multi-level question annotations and introduces an evaluation of active perception capabilities.
    \item We introduce ZoomEarth, a novel active perception framework that adaptively uses the cropping-zooming tool, and propose a Region-Guided reward tailored to the unique spatial properties of RS images.
    \item Our method achieves state-of-the-art (SOTA) on LRS-GRO and three public UHR RS datasets. Furthermore, it demonstrates strong potential as a foundation model for UHR RS agents, integrating with downstream tasks like cloud removal, segmentation, and image editing in a training-free setting.

\end{itemize}



%% file: sec/2_relatedwork.tex
\section{Related Work}
\label{sec:RelatedWork}
\subsection{High-Resolution Image Understanding}
High-resolution image understanding persists one of the key challenges for Large Vision-Language Models (LVLMs)~\cite{shi2025scalingvisionpretraining4k, li2024flexattentionefficienthighresolutionvisionlanguage, li2024visualrwkvhduhdadvancinghighresolution}. Some studies on LVLMs~\cite{kuckreja2023geochat,10817639} propose using positional embedding interpolation to increase input resolution. However, when the input image size is further enlarged, the risk of excessive downsampling remains. Recently, advanced models have introduced dynamic-resolution strategies~\cite{bai2025qwen25vltechnicalreport, wu2024deepseekvl2mixtureofexpertsvisionlanguagemodels, chen2025expandingperformanceboundariesopensource, liang2025pyramidtokenpruninghighresolution}, enabling native understanding of high-resolution images. Nevertheless, such approaches impose additional processing overhead and are susceptible to redundant visual information~\cite{li2025idalignropeconsciouspositionremapping, zhang2025hrscenefarvlmseffective}. In UHR RS imagery, small objects are significantly more numerous and densely distributed~\cite{sun2024ultrahighresolutionsegmentationboundaryenhanced, yi2025globallocalcrossattentionnetworkultrahigh}, further highlighting this challenge.
Some studies~\cite{luo2024feasteyesmixtureofresolutionadaptation, li2024monkeyimageresolutiontext, xu2024llavauhdlmmperceivingaspect} attempt to introduce separate pathways to preprocess high-resolution image features. LRS-VQA~\cite{luo2025largevisionlanguagemodelmeets} employs a Dynamic Image Pyramid to progressively crop local ROIs based on attention maps. GeoLLava-8k~\cite{elgendy2024geollava} adopts an adaptive token clustering strategy to remove background tokens. However, these models rely on manually predefined external rules and can effectively reduce redundant visual tokens only in specific scenarios (\eg, bridge inspection with large waterbody backgrounds~\cite{li2024learning}), thereby lacking applicability in diverse VQA contexts.
In this paper, we introduce the concept of active perception, enabling the VLM to obtain global context from entire RS imagery and autonomously acquire  fine-grained local features, thereby enhancing its comprehension accuracy and reasoning efficiency.

\subsection{Visual Reasoning through COT}

With the widespread application of CoT~\cite{Huang_2025, chen2025scalingrllongvideos, li2025unimumer}, VLMs have enhanced their visual perception abilities through reasoning, such as grounding tasks in images ~\cite{liu2025visualrftvisualreinforcementfinetuning, shao2024visualcotadvancingmultimodal}. However, reasoning in the textual modality can only rely on single-pass image perception, which entails the risk of missing critical visual information~\cite{tong2024eyeswideshutexploring}, leading to recent methods ~\cite{su2025pixelreasonerincentivizingpixelspace, wu2023vguidedvisualsearch, zheng2025deepeyesincentivizingthinkingimages} that attempt to expand the perceptual boundaries of VLMs by leveraging external tools.
For instance, \citet{hu2024visualsketchpadsketchingvisual} grants VLMs the ability to use tools for manipulating images, enabling them to enhance reasoning in mathematical and visual tasks by drawing auxiliary lines. \citet{huang2025visualtoolagentvistareinforcementlearning} finds that models trained with reinforcement learning (RL) can effectively use appropriate tools for complex multimodal reasoning. Groundlight AI~\citep{kumar2025reinforcingvlmsusetools} introduces GRPO training, teaching model to use zoom-in tools for reasoning, which improves the ability to recognize small objects in natural images.
These studies demonstrate that visual reasoning with tool invocation can significantly enhance the reasoning abilities of VLMs in complex tasks. However, VLMs with tools~\citep{su2025pixelreasonerincentivizingpixelspace, zheng2025deepeyesincentivizingthinkingimages} mainly rely on OCR and object detection to construct supervised data, which limits their applicability to RS imagery that lacks textual elements and contains densely distributed small objects. In this work, we introduce a localization process with varying region scales and a Region-Guided reward to enhance the model's capability for regional perception in RS images.



%% file: sec/3_method.tex
\begin{figure*}
    \centering
    \includegraphics[width=0.96\linewidth]{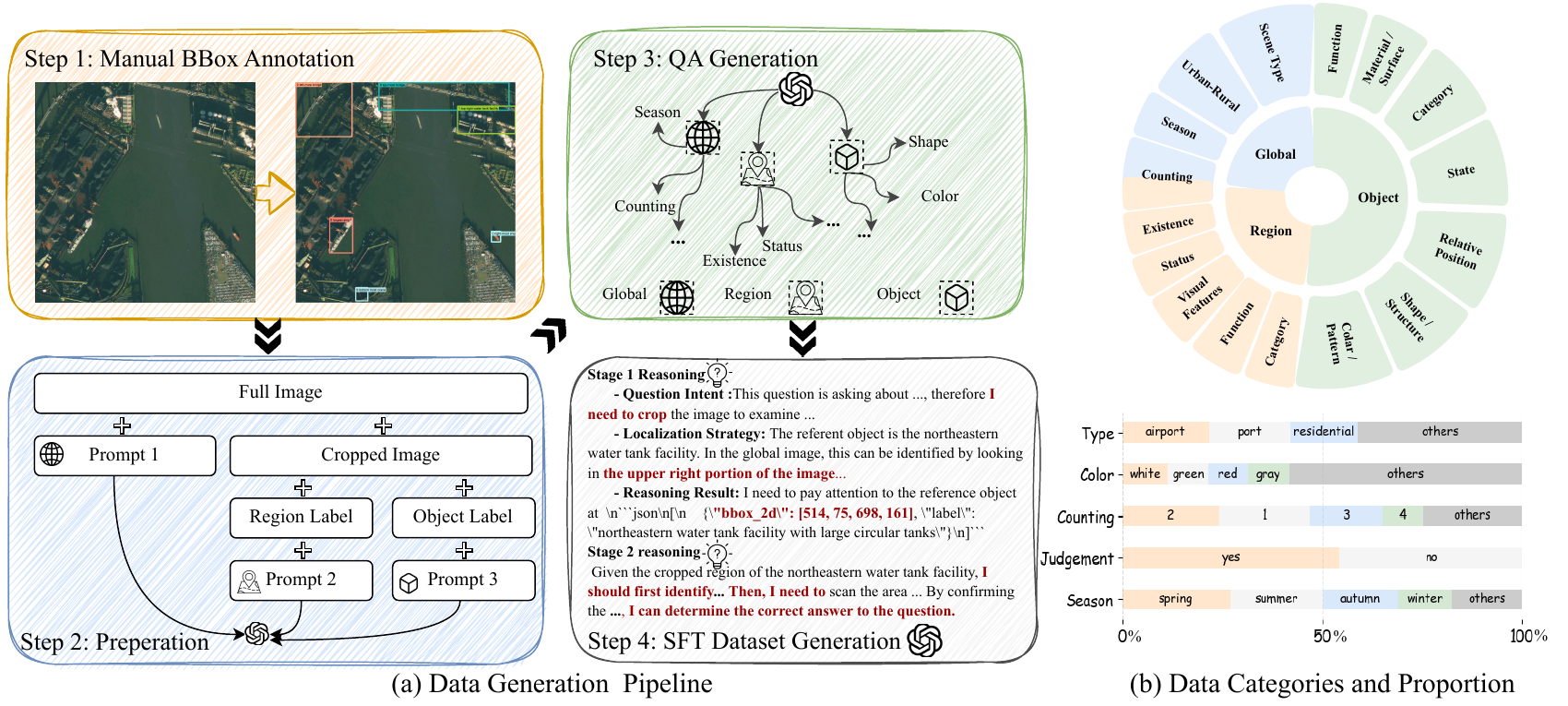}
    \vspace{-3mm}
    \caption{(a) Construction pipeline of our proposed LRS-GRO dataset, in which manual filtering and refinement are performed after Step 3. (b) The upper chart shows the 17 question types in the LRS-GRO dataset, whereas the lower chart shows the distribution of typical answer categories, demonstrating the dataset's balance.}
    \label{data}
    \vspace{-5mm}
\end{figure*}
\section{Method}

\subsection{Active Perception Oriented RS Dataset}

Although recent studies~\cite{10.1145/3746027.3758235, adejumo2025visioncentricremotesensing, wang2024earthvqa} have emphasized the importance of benchmark datasets, resources specifically designed for high-resolution RS images remain scarce. Conventional RS VQA tasks typically operate on images of moderate resolution where objects are readily identifiable. In contrast, the UHR RS image, with its vast pixel space and sparse target distributions, poses a substantial challenge for direct global observation. 
To address these challenges, we first propose a new benchmark task, Active Perception Oriented (APO) RS VQA, which requires models to actively locate, discriminate, and reason over fine-grained ROIs.

\begin{definition}[Active Perception Oriented  RS VQA]
The \textbf{APO RS VQA} task evaluates two core capabilities:  
(i) accurately answering questions about UHR RS images;  
(ii) identifying and attending to image regions that are relevant to the given question.
\end{definition}

This task involves an iterative active perception process and requires a comprehensive evaluation of a model's semantic reasoning and spatial attention, assessing not only answer accuracy but also its active perception capability through ROI-based IoU.




\subsubsection{Data Annotation Pipeline}

To support this new task, we construct the LRS-GRO benchmark dataset using a novel semi-automatic annotation pipeline. Although LRS-VQA~\cite{luo2025largevisionlanguagemodelmeets} has proposed a fully automated annotation method, it suffers from severe hallucinations of GPT when dealing with UHR RS images, making it unable to provide highly accurate BBox labels. To address this limitation, our process begins with manual annotation of BBoxes and their corresponding categories based on unique spatial characteristics (\eg, ``the top-most bridge'' and ``the building with a yellow roof''). Moreover, we explicitly distinguish between region-level and object-level BBoxes, providing hierarchical labels that better capture the unique properties of RS images. The Region-level BBoxes focus on planned land areas composed of multiple objects with distinct semantic labels, such as airports, residential communities, and industrial zones, and thus contain finer-grained compositional elements. In contrast, object-level BBoxes represent semantically homogeneous single objects, such as houses, airplanes, and ships.


Based on this, as shown in Fig.~\ref{data}, we submit the entire image, cropped regions, and object-based enlarged crops to GPT-4o~\cite{openai2024gpt4ocard} to generate candidate Question–Answer (QA) pairs concerning attributes such as the target's color, shape, and status. In total, 40k candidate QA pairs are generated, which are then manually filtered and refined to retain the highest-quality 13k pairs while maintaining answer balance. For constructing the SFT dataset, we also instruct GPT-4o to produce step-by-step reasoning processes that incorporate the cropping procedure.



\subsubsection{Dataset Composition and Structure}

Following LRS-VQA~\cite{luo2025largevisionlanguagemodelmeets}, our LRS-GRO benchmark is curated from imagery sourced from FAIR1M-1.0~\citep{sun2021fair1mbenchmarkdatasetfinegrained}, GLH-Bridge~\citep{li2024learning}, and STAR~\citep{li2024scene}. The final dataset comprises 1,224 high-resolution images (4,000–5,000 px), 3,592 bounding boxes, and 13,245 questions. Among them, 1,000 samples with detailed step-by-step CoT annotations constitute the SFT dataset.

We define 17 major question categories organized into 3 hierarchical spatial levels: 

\begin{itemize}
    \item \textbf{Global:} Counting, Season, Urban–Rural, Scene Type
    \item \textbf{Region:} Counting, Existence, Status, Visual Features, Function, Category
    \item \textbf{Object:} Function, Material / Surface, Category, State, Relative Position, Shape / Structure, Color / Pattern
\end{itemize}

For region and object-level tasks, we provide the reference BBoxes to facilitate supervision and evaluation. This hierarchical structure is a key design choice, as it enables models to adaptively determine whether cropping is necessary based on the question's scope. Table~\ref{dataset} compares our dataset with existing benchmarks, underscoring its competitiveness in terms of resolution, scale, and its distinctive emphasis on active perception oriented VQA.  More details are provided in Appendix~6.

\begin{table}[t]
\centering
\caption{Comparison between RS vision-language benchmarks and our proposed dataset. “MME-R-W” denotes MME-Real-World~\citep{zhang2024mme}. The symbols \cmark, \pmark, and \xmark represent manual, semi-automatic, and fully automatic annotation methods, respectively. Additional human verification is excluded.}
\vspace{-2mm}
\scalebox{0.66}{
\begin{tabular}{l c c c c}
\toprule
Dataset & Avg. Resolution & Active Perception& \multicolumn{2}{c}{VQA} \\
 \cmidrule(lr){4-5}
 & & & Volume & Annotation \\
\midrule
RSVQA~\cite{RSVQA}  & 512$\times$512 & N& 111,134 & \xmark \\
RSVQA-HR ~\cite{RSVQA}  & 1,024$\times$1,024 & N& 1,066,316 & \xmark \\
VRSBench~\citep{li2024vrsbenchversatilevisionlanguagebenchmark}  & 512$\times$512 & N & 123,231 & \xmark \\
RSIEval~\cite{hu2025rsgpt}  & 512$\times$512 & N & 933 & \cmark \\
XLRS-Bench~\citep{wang2025xlrsbenchmultimodalllmsunderstand} & 8,500$\times$8,500 & N& 32,389 & \cmark \\
MME-R-W~\citep{zhang2024mme}  & 2000$\times$1500 & N& 29429 & \cmark \\
LRS-VQA~\citep{luo2025largevisionlanguagemodelmeets} & 5,000$\times$5,000 & N & 7333& \xmark \\
\midrule
LRS-GRO & 5,000$\times$5,000 & Y & 13245 & \pmark \\
\bottomrule
\end{tabular}
}
\vspace{-6mm}
\label{dataset}
\end{table}

\subsection{ZoomEarth}

Building on the LRS-GRO benchmark, we propose ZoomEarth, an adaptive cropping-zooming framework designed for active perception. The core idea is to mimic human-like visual search by first gaining a holistic understanding from a coarse overview, then actively focusing on specific regions for detailed inspection. As shown in Fig.~\ref{flowchart}, ZoomEarth first processes a downsampled version of the entire UHR image to understand the global context. Subsequently, for questions requiring fine-grained detail, the model actively predicts a BBox for the relevant ROI, which is then cropped from the original high-resolution image and re-fed into the model for a detailed perception. To learn this complex behavior, we employ a two-stage training strategy.

\subsection{Training via SFT and RL}


\subsubsection{Stage 1: Supervised Fine-Tuning}

We first initialize the model through SFT. In this stage, the Qwen2.5-VL-3B~\cite{bai2025qwen25vltechnicalreport} model is fine-tuned on our SFT dataset. This process enables the model to focus on RS domain knowledge, learn to follow the predefined formats for tool invocation. In addition, it enables the model to build a fundamental capability to distinguish among task levels and their respective localization requirements.



\subsubsection{Stage 2: Reinforcement Learning}

While SFT teaches imitation, we employ reinforcement learning to build a more robust and generalizable decision-making policy. We adapt GRPO~\cite{shao2024deepseekmathpushinglimitsmathematical} for this stage. Its critic-free design offers significant memory efficiency, making it particularly well-suited for handling the long visual token sequences generated from UHR images.


\subsubsection{Reward Design for Active Perception}

A significant challenge in the RL stage is designing an effective reward function. Previous research~\cite{liu2025visualrftvisualreinforcementfinetuning,jiang2025vlmr3regionrecognitionreasoning,shen2025vlmr1stablegeneralizabler1style} on VLMs trained with RL often employ IoU as the primary reward. However, due to the foundational weakness of VLMs on UHR images, predicted BBoxes often deviate significantly from the GT, causing the IoU reward to remain zero and provide no learning signal as shown in Fig.~\ref{reward}.

To address this issue, we propose the Region-Guided reward, specifically designed for UHR RS images. This reward is inspired by the observation that geographic objects often exhibit strong spatial associations, such as aircraft located near airport terminals. Therefore, we argue that predictions located near the GT BBox should be rewarded. The Region-Guided reward provides a dense, fine-grained guidance signal by rewarding predictions based on their distance to the GT, as formulated below:

\begin{equation}
    r_{R-G}=sigmoid(\frac{\alpha}{distance+\epsilon})
\end{equation}
where $\alpha$ denotes the scale number related to the resolution and  $\text{distance}$ denotes the Euclidean distance between the center of the predicted BBox and the GT BBox center, and $\epsilon$ is a small constant to prevent numerical overflow. The final reward is defined as:

\begin{figure}[t]
    \centering
    \includegraphics[width=1.0\linewidth]{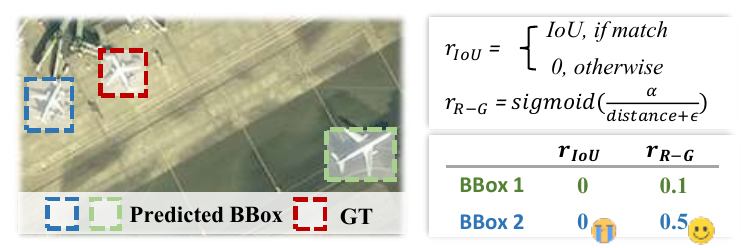}
    \vspace{-6mm}
    \caption{Comparison between $r_{IoU}$ and $r_{R-G}$.}
    \label{reward}
    \vspace{-6mm}
\end{figure}

\begin{equation}
    Reward=r_{IoU}+r_{R-G}+r_{answer}+\beta r_{pattern}
\end{equation}
where $r_{\text{IoU}}$ represents the IoU reward, the same as in VLM-$R^3$~\cite{jiang2025vlmr3regionrecognitionreasoning},  $r_{\text{answer}}$ is the normalized answer reward derived from word similarity~\cite{10.1145/219717.219748}, and $r_{\text{pattern}}$ denotes the reward for the output format, assigned a value of 1 if correct and 0 if incorrect with the coefficient $\beta=0.05$ controlling its influence. More explanations are provided in Appendix~7. 


 

\subsection{Toolkit for Downstream Tasks}
Through the preceding training stages, the model acquires the capability to adaptively crop ROIs, ranging from region-level to object-level. This introduces the potential of a foundation model tailored for a wide spectrum of downstream tasks in the RS domain. 
Considering that directly performing tasks such as denoising and clouding removal on high-resolution global images incurs substantial computational costs and often yields limited practical benefits. As the framework shown in Fig.~\ref{down}, our cropping–zooming method enables the model to more efficiently focus on user-specified regions of interest in UHR RS. In a training-free manner, we integrate multiple advanced downstream models and design task-specific instructions to achieve adaptive tool invocation. Based on the proposed ZoomEarth model, we construct a scalable RS agent capable of performing functions including segmentation, denoising, clouding removal, and image editing.


%% file: sec/4_experiment.tex
\section{Experiment}
\subsection{Settings and Implementation Details}
\subsubsection{Dataset}
\textbf{Training Data.} We utilize the LRS-GRO dataset, which is split into approximately 3,500 samples for training and 10,000 samples for testing.  
In the first stage, we fine-tune the model on the SFT dataset, with tool-necessary and tool-unnecessary subsets in a 2:1 ratio. In the second stage, we fine-tune the model on 2,500 samples using RL to enhance its tool utilization and generalization ability.

\paragraph{\textbf{Test Data.}} For evaluation, we assess model performance on the LRS-GRO test set and  other datasets with resolutions exceeding $5000 \times 5000$, including {MME-Real-World}~\cite{zhang2024mme}, {XLRS-Bench}~\cite{wang2025xlrsbenchmultimodalllmsunderstand} and {GeoLLava-8k}~\cite{wang2025geollava8kscalingremotesensingmultimodal}. 


\subsubsection{Evaluation Metrics}
\textbf{Accuracy (Acc).} We report Average Accuracy (Avg. Acc) for the VQA task across the aforementioned datasets. For LRS-GRO, the correctness of predictions is evaluated by measuring the semantic similarity between the predicted and ground-truth answers using WordNet~\cite{10.1145/219717.219748}, where a similarity score higher than 0.8 is regarded as correct\cite{xiao2021next, Hudson_2019_CVPR, luo2025largevisionlanguagemodelmeets}. For MME-Real-World~\cite{zhang2024mme}, XLRS-Bench~\cite{wang2025xlrsbenchmultimodalllmsunderstand}, we follow the evaluation setting provided by their respective datasets. Due to the lack of UHR RS datasets, we additionally incorporate the training dataset proposed by GeoLLava-8k\cite{wang2025geollava8kscalingremotesensingmultimodal} for testing, applying the same evaluation settings as that used for our proposed LRS-GRO dataset.

\paragraph{\textbf{APO IoU.}} 
Standard VQA accuracy is insufficient as it does not evaluate how the model arrived at an answer. To measure the effectiveness of the localization step, we introduce the active perception oriented (APO) IoU. This metric is defined as the IoU between the predicted ROI bounding box and the ROI annotation. The APO IoU metric is applied to the region and object-level questions, as these tasks are designated to require tool invocation. A critical component of this evaluation is that if a model fails to generate a valid bounding box for a required task, its APO IoU for that sample is automatically set to 0. This directly penalizes failures in the active perception process, providing a comprehensive assessment of the model's tool-use capability.

\subsubsection{Implementation Details}
All experiments are conducted on 8 NVIDIA A800 GPUs.  
During training, the images are processed at an input resolution of 512 pixels (before cropping). 
In the RL stage, the sampling temperature is set to 0.7, while for all evaluation experiments, the temperature is fixed at 0.01. Regarding the learning rate, it is set to 3e-5 in the SFT stage and 1e-7 in the RL stage.
More details are provided in Appendix~8.

\begin{table*}[tbp]
\centering
\caption{Experimental results of various models on LRS-GRO. ``Max Size'' refers to the maximum resolution that the model can process as input. “method*” indicates we reproduce the SFT stage of existing methods using the same training data as ours. ``-'' denotes that the model lacks ROI localization outputs, making APO IoU evaluation impossible.}
\vspace{-2mm}
\scalebox{1.0}{
\begin{tabular}{lccccccc}
\hline
\hline
\textbf{Leaderboard}  & \textbf{LLM} & \textbf{Max Size} & \textbf{Global}& \textbf{Region}& \textbf{Object}&\textbf{Avg. Acc}   &\textbf{APO IoU}\\
LLaVA-OV-1.5~\citep{LLaVA-OneVision-1.5} & Qwen3-7B       & 2,304$\times$2,304 & 62.43 & 36.18 & 39.92 &45.33  &-\\
IXC-2.5~\citep{zhang2024internlmxcomposer25versatilelargevision}& InternLM2-7B   & 4,096$\times$4,096 & 62.43 & 40.00 & 47.32 & 50.00  &-\\
InternVL3~\citep{wang2025internvl35advancingopensourcemultimodal}&  InternLM3-8B & 3200$\times$3200   & \textbf{71.60} & 44.58 & 47.80 &53.67  &-\\
Geochat~\citep{kuckreja2023geochat}& Vicuna-1.5-7B  & 504$\times$504     & 62.43 & 36.79 & 40.72 &45.88   &-\\

\multirow{4}{*}{Qwen2.5-VL~\citep{bai2025qwen25vltechnicalreport}} & \multirow{2}{*}{Qwen2.5-7B} & 1024$\times$1024 & 69.39 & 38.47 & 43.58 &49.62  &-\\
                            &                             & 3333$\times$3333 & 70.06 & 39.08 & 46.43 &51.43  &-\\
                            & \multirow{2}{*}{Qwen2.5-3B} & 1024$\times$1024 & 59.01 & 31.91 & 37.46 &42.25  &-\\
                            &                             & 3333$\times$3333 & 58.90 & 31.76 & 38.66 &42.83  &-\\
VLM-$R^3$~\citep{jiang2025vlmr3regionrecognitionreasoning} (w/ tools)& Qwen3-7B             & 512$\times$512                  &69.72 &44.83 &37.40 & 50.17 &19.93\\
\midrule

Geochat* & Vicuna-1.5-7B  & 504$\times$504     & 58.78 & 37.25 & 42.83 &46.09   &-\\
\rowcolor{gray!10}
ZoomEarth (ours)                        & Qwen2.5-3B                  & 512$\times$512& 63.09& \textbf{46.11}& \textbf{51.80}&\textbf{53.76}  &\textbf{34.39}\\
\hline
\hline
\end{tabular}
}
\vspace{-1mm}
\label{t1}
\end{table*}

\begin{table*}[tbp]
\centering
\caption{Zero-shot comparison of various models on MME-RealWorld-RS~\citep{zhang2024mme}, XLRS-bench~\citep{wang2025xlrsbenchmultimodalllmsunderstand}, GeoLLava-8k~\citep{wang2025geollava8kscalingremotesensingmultimodal}. “–” indicates unavailable results due to potential data leakage risk from overlapping GeoLLava-8k and GeoChat training data.}
\vspace{-2mm}
\scalebox{1.0}{
\begin{tabular}{lccccc}
\hline
\hline
\textbf{Leaderboard}  & \textbf{LLM} & \textbf{MME-realworld-RS}& \textbf{XLRS-bench}& \textbf{GeoLLava-8k}& \textbf{Avg. Acc} \\
LLaVA-OV-1.5~\citep{LLaVA-OneVision-1.5} & Qwen3-7B       & 33.20& 39.60 & 31.48& 36.40\\
IXC-2.5~\citep{zhang2024internlmxcomposer25versatilelargevision}& InternLM2-7B   & 36.12& 31.50 & 36.20 & 34.61 \\
InternVL3-8B~\citep{wang2025internvl35advancingopensourcemultimodal}& InternLM3-8B   & 41.00 & 36.70 &37.60  &38.43 \\
Geochat~\citep{kuckreja2023geochat}& Vicuna-1.5-7B  & 28.62& 22.03& - & -\\
Qwen2.5-VL~\citep{bai2025qwen25vltechnicalreport}& Qwen2.5-3B& 23.94& 36.00& 34.64& 29.97\\
VLM-$R^3$~\citep{jiang2025vlmr3regionrecognitionreasoning} (w/ tools)& Qwen3-7B   &39.80 &39.10 &34.74 &37.88 \\
\rowcolor{gray!10}
ZoomEarth (ours)   & Qwen2.5-3B& \textbf{44.10}&  \textbf{40.20}& \textbf{38.61}&\textbf{40.97} \\
\hline
\hline
\end{tabular}
}
\label{t2}
\vspace{-4mm}
\end{table*}

\subsection{Main Results}

Our model, built upon the Qwen2.5-VL 3B~\cite{bai2025qwen25vltechnicalreport} architecture, consistently outperforms multiple baseline models on the LRS-GRO dataset, as shown in Table~\ref{t1}.
Compared with InternVL3-8B that supports higher-resolution inputs, our model, with an initial input resolution of only 512, achieves substantial advantages in region-level and object-level tasks, improving accuracy by 1.53\%  and 4.00\%.
For global-level tasks that  emphasize overall context rather than regional understanding, ZoomEarth is predictably lower than some larger models but still achieves a notable advantage over models of comparable size, such as Qwen2.5-VL-3B~\cite{bai2025qwen25vltechnicalreport}.
To ensure fairness, we adopt the Geochat~\citep{kuckreja2023geochat} approach and fine-tuned its pre-trained weights on LRS-GRO using the same data. We observe that, despite Geochat having been trained on a large amount of additional RS data, ZoomEarth still demonstrates a significant advantage.
Compared with the publicly available general-purpose tool-calling model VLM-$R^3$~\cite{jiang2025vlmr3regionrecognitionreasoning}, our smaller model achieves a 3.59\% performance gain. Moreover, our {APO IoU} reaches {34.39\%}, which is substantially higher than VLM-$R^3$~\cite{jiang2025vlmr3regionrecognitionreasoning}. This highlights  the effectiveness and significance of our proposed dataset and training method for RS applications.

To evaluate the generalization capability of our model trained on the LRS-GRO dataset, we conduct zero-shot testing on three UHR RS datasets 
with all models evaluated under an input resolution of 1024.
As shown in Table~\ref{t2}, compared with various VLMs from both general and RS domains, our model achieves SOTA performance on MME-RealWorld-RS~\cite{zhang2024mme}, GeoLLava-8k~\cite{wang2025geollava8kscalingremotesensingmultimodal}, and XLRS-Bench~\cite{wang2025xlrsbenchmultimodalllmsunderstand}.
Specifically, XLRS-Bench~\cite{wang2025xlrsbenchmultimodalllmsunderstand} provides predefined BBoxes in questions for region localization. Although this differs from our proposed framework that the model actively generates the BBox of the ROI, our method still demonstrates strong generalization ability.

\subsection{Ablation Study}

\paragraph{\textbf{Ablation on cropping.}}
To assess the effectiveness of the cropping–zooming tool, we conducted an ablation study by removing its application process. Specifically, we trained the model on the same dataset but required it to produce final answers without access to the cropped images. As shown in Table~\ref{a1}, the proposed active perception strategy yielded an average improvement of 4.39\% across four datasets. In addition, experiments conducted on other UHR RS datasets are under the zero-shot setting, which highlights the remarkable generalization ability of ZoomEarth.

\begin{table}[t]
\centering
\caption{Ablation on cropping. ``MME-R-W'' means MME-Real-World~\citep{zhang2024mme}, XLRS refers to XLRS-bench~\citep{wang2025xlrsbenchmultimodalllmsunderstand}. `+ Cropping' indicates the incorporation of cropping–zooming functionalities.}
\vspace{-2mm}
\scalebox{0.75}{
\begin{tabular}{lcccc}
\toprule
\textbf{Methods}   & \textbf{LRS-GRO} & \textbf{MME-R-W} & \textbf{XLRS} & \textbf{GeoLLaVA-8k} \\
\midrule
ZoomEarth & 51.10& 42.10& 30.00 & 34.30 \\
\rowcolor{gray!10}
+ Cropping & 53.64{\color{darkgreen}\scriptsize$\uparrow$2.57}& 44.10{\color{darkgreen}\scriptsize$\uparrow$2.00}& 40.20{\color{darkgreen}\scriptsize$\uparrow$10.20} & 38.61{\color{darkgreen}\scriptsize$\uparrow$4.31} \\
\bottomrule
\end{tabular}
}
\label{a1}
\vspace{-3mm}
\end{table}

\paragraph{\textbf{Ablation on RL stage.}}
To investigate the impact of the two-stage training strategy on enhancing the model's tool reasoning capability, we removed the RL stage on the LRS-GRO dataset. An ablation study was then conducted on the MME-RealWorld-RS dataset to avoid overfitting to the training set (LRS-GRO). As shown in Table~\ref{a2}, after SFT alone, invoking the cropping tool unexpectedly resulted in a performance drop. This finding suggests that SFT-based imitation learning primarily aids the model in acquiring output formats, whereas RL is crucial to the model's  reasoning with tools and active perception capabilities.

\begin{table}[t]
\centering
\caption{Impact of RL on reasoning performance with tool usage. }
\vspace{-2mm}
\scalebox{0.89}{
\begin{tabular}{lcccc}
\toprule
\textbf{Methods}   & \textbf{Color}& \textbf{Count }& \textbf{Position}&\textbf{Overall}\\
\midrule
SFT& 45.71& 29.78& 51.83&42.80\\
\rowcolor{gray!10}
+ Cropping & 45.40{\color{red}\scriptsize$\downarrow$0.31}& 27.90{\color{red}\scriptsize$\downarrow$1.88}& 49.58{\color{red}\scriptsize$\downarrow$2.25}&41.30{\color{red}\scriptsize$\downarrow$1.50}\\
\midrule
RL& 48.47& 29.47& 47.61&42.10\\
\rowcolor{gray!10}
+ Cropping & 49.69{\color{darkgreen}\scriptsize$\uparrow$1.22}& 31.66{\color{darkgreen}\scriptsize$\uparrow$2.19}& 50.14{\color{darkgreen}\scriptsize$\uparrow$2.53}&44.10{\color{darkgreen}\scriptsize$\uparrow$2.00}\\
\bottomrule
\end{tabular}
}

\label{a2}
\vspace{-3mm}
\end{table}

\begin{table}[t]
\centering
\caption{Ablation results for reward based on LRS-GRO.}
\vspace{-2mm}
\scalebox{0.95}{
\begin{tabular}{llcccc}
\toprule
 \textbf{$r_{R-G}$}
&\textbf{$r_{IOU}$} 
& \textbf{Global}& \textbf{Region}& \textbf{Object}&\textbf{Avg. Acc}\\
\midrule
 \ding{55}
&\ding{55} 
& 62.54& 43.36& 51.06&52.67\\
 \ding{51}
&\ding{55} 
& 62.21& 45.80& 50.83&52.94\\
 \ding{55} 
& \ding{51}
& 62.43& 43.97& 51.11&52.79\\
\rowcolor{gray!10}
 \ding{51}
&\ding{51}
& \textbf{63.09}& \textbf{46.11}& \textbf{51.80}&\textbf{53.76}\\
\hline
\end{tabular}
}
\label{a3}
\vspace{-6mm}
\end{table}

\begin{figure*}[t]
    \centering
    \includegraphics[width=0.92\linewidth]{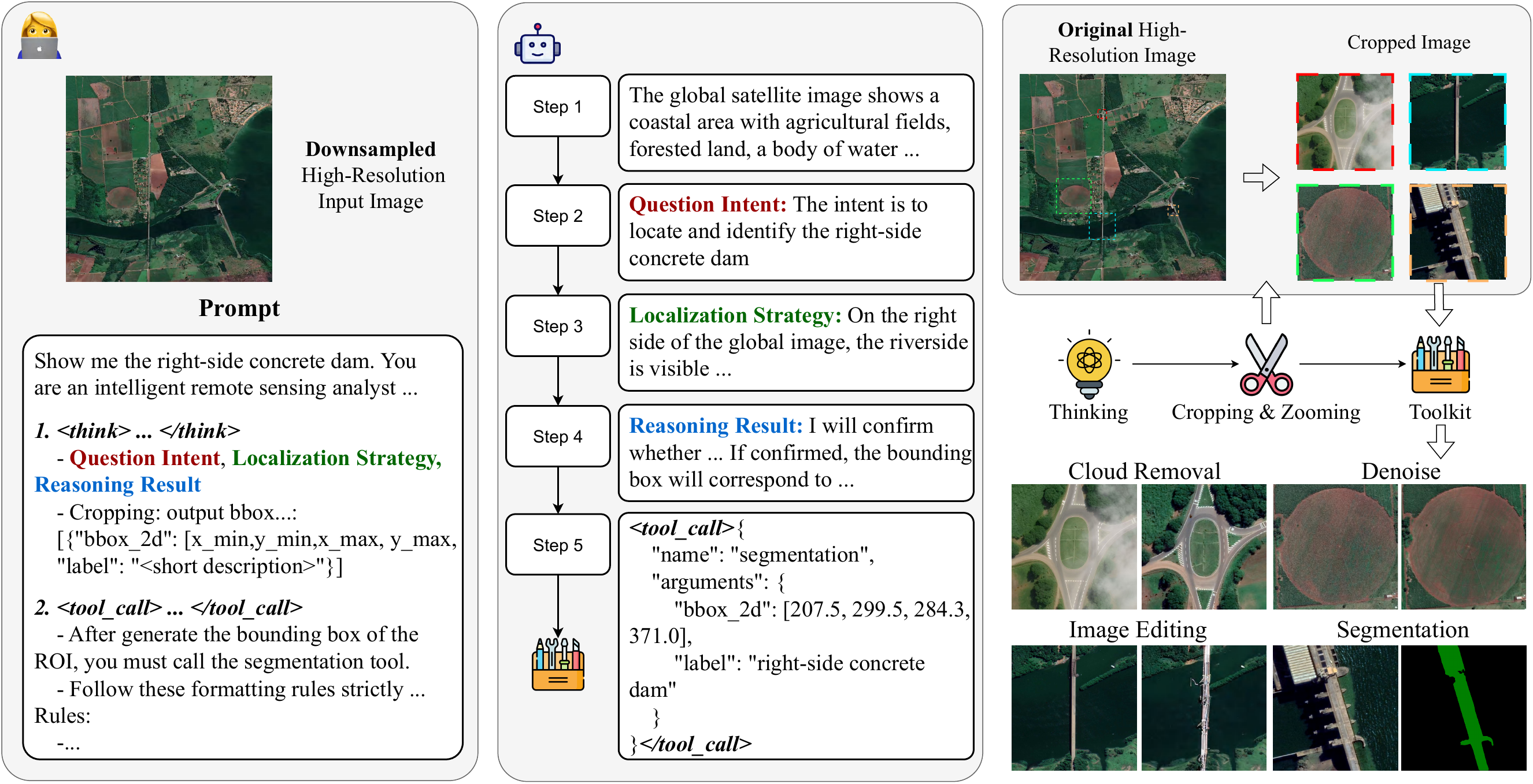}
    \vspace{-2mm}
    \caption{ZoomEarth inference pipeline with open tool invocation for cloud removal, denoising, image editing, and semantic segmentation. Due to the unavailability of suitable data for cloud removal, we simulated cloud-covered images using an image editing model~\citep{openai2024gpt4ocard}.}
    \label{down}
    \vspace{-5mm}
\end{figure*}

\paragraph{\textbf{Ablation on specific rewards.}}
We conducted an ablation study on the LRS-GRO dataset to examine the effects of different rewards, focusing on the IOU reward and the Region-Guided reward. As shown in Table~\ref{a3}, removing the Region-Guided reward resulted in a performance drop of 0.97\%, which is greater than the 0.82\% drop observed when removing the IOU reward. This finding suggests that, for UHR RS images, the Region-Guided reward plays a more significant role than the IOU reward. Notably, compared with the setting without either IOU or Region-Guided rewards, adding the Region-Guided reward improved performance on region QA tasks by 2.44\%, demonstrating stronger adaptability to  region-level understanding  questions.

\paragraph{\textbf{Ablation on different resolution input.}}
As shown in Table~\ref{a4}, we examine the performance differences of the model under varying input resolutions. The results indicate that, although increasing the resolution of input images provides more fine-grained visual information, it does not yield significant performance improvements, which means the model is easily disturbed by the redundant information introduced by a large number of visual tokens. However, after invoking the perception tool, the model still achieves a performance gain of over 2\% at an input resolution of 512, suggesting that the our approach can effectively help the model overcome its performance bottleneck when processing high-resolution images and offers a significant speed advantage compared to $3333 \times 3333$ resolution input.

\begin{table}[t]
\centering
\caption{Ablation on inputting resolution. The model is evaluated during inference across various input resolutions under identical training data conditions.}
\vspace{-2mm}
\scalebox{0.68}{
\begin{tabular}{ccccccc}
\toprule
 \textbf{Cropping}&\textbf{Max Size}& \textbf{Global}& \textbf{Region}& \textbf{Object}&\textbf{Avg. Acc} &\textbf{Speed (it/s)}\\
 \ding{55}& 512$\times$512& 64.31& 42.14& 47.17 &50.86 &3.69\\
 \ding{55}&1024$\times$1024& 63.76 & 41.37 & 48.32 &51.16 &3.50\\
 \ding{55}&3333$\times$3333& 63.31 & 40.61 & 48.37 &50.92 &2.04\\
 \rowcolor{gray!10}
 \ding{51}&512$\times$512& \textbf{64.53}& \textbf{44.43}& \textbf{51.06}&\textbf{53.43} &2.38\\
 \hline
\end{tabular}
}
\label{a4}
\vspace{-6mm}
\end{table}

\subsection{Unified Framework for Downstream Tasks}

In practical applications, users typically focus on specific regions within high-resolution RS images, necessitating that the model possess the capability to autonomously identify and process the target regions.
In our proposed approach, the model can identify ROIs using the cropping-zooming tool, and then performs subsequent processing reasoning. Through prompt-based instructions, we enable the model to invoke a broader set of tools and integrate specialized models from multiple domains, including cloud removal~\cite{liu2025effectivecloudremovalremote}, denoising~\cite{Li_Zhang_Luo_Meng_2025}, image editing~\cite{liu2025text2earthunlockingtextdrivenremote}, and segmentation~\cite{li2025segearthOV}, as illustrated in Fig.~\ref{down}. In implementation, we specify in the instructions that the model should output the BBox of the ROI within the tool invocation section, crop it from the original high-resolution image, and route it to the appropriate downstream model based on the determined tool name.
Although our experiments are limited to a training-free setting, the results demonstrate that the model can adaptively select suitable tools from the toolkit according to the task requirements. The ZoomEarth model shows strong potential for UHR RS image processing and can be extended to a wide range of downstream tasks.

%% file: sec/5_conclusion.tex
\section{Conclusion}
In this work, we introduce the concept of active perception to tackle the challenges of UHR RS VQA. We present LRS-GRO, a high-quality and diverse benchmark dataset spanning global, region, and object-level questions, enabling the stimulation and evaluation of active perception capabilities. Our proposed ZoomEarth framework adaptively calls the cropping–zooming tool to actively explore and extract fine-grained information from UHR images, guided by our proposed reward mechanism tailored to the distributional characteristics of RS scenes. Experimental results demonstrate competitive performance, strong generalization, and a scalable foundation for integrating downstream tasks. 
We anticipate that ZoomEarth will form the foundation for designing future autonomous RS agents, equipping them with the fundamental capability to actively seek information within complex, large-scale environments.




%% file: sec/X_suppl.tex
\clearpage
\setcounter{page}{1}
\maketitlesupplementary

\section{LRS-GRO Dataset}
\label{appendix:6}
\subsection{Data Generation Details}

The dataset construction process involves three distinct settings: \textbf{object-level}, \textbf{region-level}, and \textbf{global-level} VQA with their corresponding CoT annotations.  
For each setting, both the input to the annotation model (GPT-4o) and the expected output format are defined explicitly.

\subsection{Region-level and Object-level VQA.}
For non-global cases, human annotators first manually select ROIs in the image by drawing bounding boxes.  
Each ROI is further categorized as either:
\begin{itemize}
    \item \textbf{Region}: a spatially extended area containing multiple objects or structures, where the cropped image is taken directly from the bounding box.
    \item \textbf{Object}: a single target object of interest, where the bounding box is expanded before cropping to provide additional surrounding context.
\end{itemize}

For each selected ROI, the following information is passed to GPT-4o:
\begin{enumerate}
    \item A downsampled version of the full image (for global context).
    \item A high-resolution cropped image derived from the bounding box (region: direct crop; object: expanded crop).
    \item The human-provided label describing the object's or region's position and type.
    \item The bounding box coordinates.
\end{enumerate}

Then GPT-4o will generate Question-Answer pairs based on prompts shown in Appendix~\ref{p1}.

\begin{tcolorbox}[title=Dataset Annotation Prompt of Object Questions,
                  colframe=black!20,
                  colback=white,
                  coltitle=black,
                  fonttitle=\bfseries,
                  breakable]
\label{p1}
\textbf{You are an expert dataset annotator for an \emph{Object-level Visual Question Answering (VQA)} dataset.}  

You will be provided with:
\begin{enumerate}[label=\arabic*.]
    \item A downsampled full image showing the global context.
    \item A high-resolution cropped image centered on the target object (ROI).
    \item A bounding box for the object: \verb|{item.get('bbox')}|.
    \item A short human-provided label describing the object's \textbf{global position and type}: \verb|{item.get('label')}|.
\end{enumerate}

\medskip
\textbf{Your tasks:}
\begin{enumerate}[label=\arabic*.]
    \item \textbf{Refine the label} into a clear and specific English name for the object, preserving its spatial description.  
    Examples:
    \begin{itemize}
        \item ``top-most sports field'' $\rightarrow$ ``top-most tennis court''
        \item ``central building'' $\rightarrow$ ``central white office building''
    \end{itemize}
    \item \textbf{Generate several high-quality, image-dependent question–answer pairs} about this object:
    \begin{itemize}
        \item Use the refined label \textbf{explicitly} (no pronouns like ``this object'' or ``it'').
        \item Be \textbf{visually grounded} — cannot be answered by common sense alone.
        \item Require looking at the provided images to answer.
        \item Avoid trivial or universal facts (\eg, airplane has wings, grass is green).
    \end{itemize}
\end{enumerate}

\medskip
\textbf{Object-level Question Categories:}
\begin{enumerate}[label=\arabic*.]
    \item \textbf{Object category refinement} — Make the label more specific based on visual cues.  
    Examples:
    \begin{itemize}
        \item ``Is the top-most sports field a tennis court or a basketball court?''
        \item ``Is the left-side bridge designed for vehicles or pedestrians?''
    \end{itemize}
    \item \textbf{Object color / pattern} — Non-obvious or detailed color info.  
    Examples:
    \begin{itemize}
        \item ``What color is the roof of the central building?''
        \item ``Is the top-right ship mainly white or blue?''
        \item ``Does the top-most tennis court have a green or red surface?''
    \end{itemize}
    \item \textbf{Object shape / structure} — Shape or structural details.  
    Examples:
    \begin{itemize}
        \item ``What is the shape of the top-most building's roof?''
        \item ``Is the left-side bridge straight or curved?''
        \item ``Is the bottom-right ship narrow or wide?''
    \end{itemize}
    \item \textbf{Object function / usage} — Purpose or role inferred from context.  
    Examples:
    \begin{itemize}
        \item ``What is the function of the central rectangular building?''
        \item ``Is the left-side bridge used for vehicles or trains?''
    \end{itemize}
    \item \textbf{Object state / motion / activity} — Current condition or movement.  
    Examples:
    \begin{itemize}
        \item ``Is the right-most vehicle moving or parked?''
        \item ``Is the top-left airplane taking off or landing?''
        \item ``Is the central crane operating or idle?''
    \end{itemize}
    \item \textbf{Object material / surface} — Visible material cues.  
    Examples:
    \begin{itemize}
        \item ``Is the left-side bridge made of metal or concrete?''
        \item ``Does the roof of the central building appear metallic or tiled?''
    \end{itemize}
    \item \textbf{Object relative position / context} — Spatial relations to nearby elements.  
    Examples:
    \begin{itemize}
        \item ``What is located beneath the top-most bridge?''
        \item ``Is there water below the bottom bridge?''
        \item ``What is on the right side of the central building?''
    \end{itemize}
\end{enumerate}

\medskip
\textbf{Guidelines:}
\begin{itemize}
    \item Use the refined label directly in all questions (no pronouns).
    \item Ensure each question is visually discriminative — answerable only by observing the images.
    \item Keep answers concise: ``yes'', ``no'', ``concrete'', ``curved'', ``green'', ``train'', ``parked'', ``water''.
    \item Each question must include:
    \begin{itemize}
        \item \verb|"category"| — one of the seven categories above.
        \item \verb|"higher_level"| — one of:
        \begin{itemize}
            \item \verb|"perception"| — visual recognition (color, shape, count)
            \item \verb|"localization"| — spatial position or relative location
            \item \verb|"attribute"| — appearance or material properties
            \item \verb|"function"| — role or purpose
            \item \verb|"reasoning"| — inferred or dynamic states
        \end{itemize}
    \end{itemize}
\end{itemize}

\medskip
\textbf{Output format:}
\begin{lstlisting}[language=json]
{
  "label": "top-most tennis court",
  "qa_pairs": [
    {
      "question": "Is the top-most sports field a tennis court or a basketball court?",
      "answer": "tennis court",
      "category": "Object category refinement",
      "higher_level": "attribute"
    },
    {
      "question": "Does the top-most tennis court have a green or red surface?",
      "answer": "green",
      "category": "Object color / pattern",
      "higher_level": "perception"
    },
    {
      "question": "Is the top-most tennis court surrounded by fences?",
      "answer": "yes",
      "category": "Object shape / structure",
      "higher_level": "reasoning"
    },
    {
      "question": "What is located next to the top-most tennis court?",
      "answer": "parking lot",
      "category": "Object relative position / context",
      "higher_level": "localization"
    }
  ]
}
\end{lstlisting}

\end{tcolorbox}

\begin{tcolorbox}[title=Dataset Annotation Prompt of Region Questions,
                  colframe=black!20,
                  colback=white,
                  coltitle=black,
                  fonttitle=\bfseries,
                  breakable]
\label{p2}
\textbf{You are an expert dataset annotator for a Visual Question Answering (VQA) dataset.}  

You will be given:
\begin{enumerate}[label=\arabic*.]
    \item A downsampled full image.
    \item A high-resolution cropped image of the region of interest (ROI).
    \item A bounding box representing the ROI: \verb|{item.get('bbox')}|.
    \item A short human-provided description of the ROI: \verb|{item.get('label')}|.
\end{enumerate}

\textbf{Your task:}
\begin{itemize}
    \item Refine the label to make it a \textbf{precise and natural English name} for this region (\eg, ``central bridge area'', ``left-most parking lot'', ``top-most construction site'').
    \item Generate several \textbf{diverse question–answer pairs} about the ROI, following the categories below.
\end{itemize}

\medskip
\textbf{Question categories (choose those applicable to the region):}

\begin{enumerate}[label=\arabic*.]
    \item \textbf{Counting} — Ask about the \textbf{number} of visible objects.  
    Examples:
    \begin{itemize}
        \item ``How many vehicles are on the bridge?''
        \item ``How many ships are docked near the pier?''
    \end{itemize}

    \item \textbf{Object existence} — Ask if certain objects are \textbf{present} in the ROI.  
    Examples:
    \begin{itemize}
        \item ``Is there a ship passing under the bridge?''
        \item ``Are there any cars in the parking lot?''
        \item ``Is any airplane on the runway?''
    \end{itemize}

    \item \textbf{Region status} — Ask about \textbf{activity, usage, or condition} of the region.  
    Examples:
    \begin{itemize}
        \item ``Is the bridge busy or empty?''
        \item ``Is the construction site still active?''
        \item ``Are there ships currently docking at the port?''
        \item ``Is the road under construction or in use?''
    \end{itemize}

    \item \textbf{Object category} — Ask about the \textbf{types of main objects} found in the ROI.  
    Examples:
    \begin{itemize}
        \item ``What types of vehicles are in the parking lot?''
        \item ``What kind of boats are docked at the pier?''
    \end{itemize}

    \item \textbf{Region function} — Ask about the \textbf{purpose or role} of the region.  
    Examples:
    \begin{itemize}
        \item ``What is this area mainly used for?''
        \item ``What is the function of this rectangular region?''
    \end{itemize}

    \item \textbf{Other visual features} — Ask about \textbf{appearance, color, or shape} of the region or its objects.  
    Examples:
    \begin{itemize}
        \item ``Are most buildings in this area red-roofed?''
        \item ``What is the overall shape of this region?''
        \item ``Is the area circular or rectangular?''
    \end{itemize}
\end{enumerate}

\medskip
\textbf{Output requirements:}
\begin{itemize}
    \item Only ask \textbf{reasonable} questions that can be answered directly from the provided images.
    \item Provide concise answers (one word or short phrase). Examples: ``yes'', ``no'', ``asphalt'', ``empty'', ``circular'', ``urban'', ``in use''.
    \item Each question must include:
    \begin{itemize}
        \item \verb|"category"|: one of the six above.
        \item \verb|"higher_level"|: one of these abstract reasoning levels:
        \begin{itemize}
            \item \verb|"perception"| — direct visual recognition
            \item \verb|"localization"| — position or spatial relation
            \item \verb|"attribute"| — appearance or measurable quality
            \item \verb|"function"| — purpose or role
            \item \verb|"reasoning"| — requires inference or contextual understanding
        \end{itemize}
    \end{itemize}
\end{itemize}

\medskip
\textbf{Output format:}
\begin{lstlisting}[language=json]
{
  "label": "top-most bridge area",
  "qa_pairs": [
    {
      "question": "How many vehicles are on the bridge?",
      "answer": "3",
      "category": "Counting",
      "higher_level": "perception"
    },
    {
      "question": "Is the bridge currently in use?",
      "answer": "yes",
      "category": "Region status",
      "higher_level": "reasoning"
    },
    {
      "question": "Are there ships passing under the bridge?",
      "answer": "no",
      "category": "Object existence",
      "higher_level": "perception"
    },
    {
      "question": "What type of vehicles are visible on the bridge?",
      "answer": "cars",
      "category": "Object category",
      "higher_level": "attribute"
    },
    {
      "question": "What is the main function of this bridge area?",
      "answer": "transportation",
      "category": "Region function",
      "higher_level": "function"
    }
  ]
}
\end{lstlisting}

\end{tcolorbox}

\subsection{Global-level VQA}
For global-level questions, the full high-resolution satellite image is directly provided to GPT-4o, together with a task-specific prompt (see Appendix~\ref{p3}).  
The model is instructed to generate scene-level, visually grounded question–answer pairs that require holistic understanding of the image, such as scene type, counting of large-scale objects, or seasonal inference.

\begin{tcolorbox}[title=Dataset Annotation Prompt of Global Questions,
                  colframe=black!20,
                  colback=white,
                  coltitle=black,
                  fonttitle=\bfseries,
                  breakable]
\label{p3}
\textbf{You are an expert dataset annotator for a Visual Question Answering (VQA) task focusing on \emph{global-level understanding} of high-resolution remote sensing images.}  

You will be provided with:
\begin{enumerate}[label=\arabic*.]
    \item A full high-resolution image covering the entire scene.
\end{enumerate}

\medskip
\textbf{Your task:}  
Generate several \textbf{high-quality, globally grounded question–answer pairs} in English about the image.

\medskip
\textbf{Requirements for each question:}
\begin{itemize}
    \item \textbf{Scene-level}, not object-level.
    \item \textbf{Visually grounded} — answerable purely by looking at the image.
    \item \textbf{Specific and unique} — avoid vague or overly general questions.
    \item Generate only if the image clearly supports it; otherwise, output an empty list.
\end{itemize}

\medskip
\textbf{Possible question types (use only when appropriate):}
\begin{itemize}
    \item \textbf{Counting} — \eg, ``How many airplanes are visible in the image?''
    \item \textbf{Urban–Rural} — \eg, ``Does this image mainly depict an urban or rural area?''
    \item \textbf{Scene Type} — \eg, ``What is the main type of area shown — airport, residential, or farmland?''
    \item \textbf{Season} — \eg, ``What season does the scene appear to be?''
\end{itemize}

\medskip
\textbf{Higher-level reasoning categories:}
\begin{itemize}
    \item \verb|perception|
    \item \verb|localization|
    \item \verb|attribute|
    \item \verb|function|
    \item \verb|reasoning|
\end{itemize}

\medskip
\textbf{Output format (strict JSON):}
\begin{lstlisting}[language=json]
{
  "qa_pairs": [
    {
      "question": "How many airplanes are visible in the image?",
      "answer": "3",
      "category": "Counting",
      "higher_level": "perception",
      "justification": "Airplanes are distinct and countable across the visible runways."
    },
    {
      "question": "What is the main type of area shown in this image airport, residential, or farmland?",
      "answer": "airport",
      "category": "Scene Type",
      "higher_level": "reasoning",
      "justification": "The image shows large runways and parked airplanes typical of an airport."
    }
  ]
}
\end{lstlisting}

\end{tcolorbox}

\subsection{SFT Dataset Annotation}
In order to train the model to master both the step-by-step reasoning process and the standardized tool invocation format for answering questions, we employed GPT-4o to generate annotations using a two-stage reasoning–cropping–reasoning CoT paradigm, as defined in the prompt provided in Appendix~\ref{p4}.

\begin{tcolorbox}[title=Reasoning-based remote sensing VQA Annotation Prompt,
                  colframe=black!20,
                  colback=white,
                  coltitle=black,
                  fonttitle=\bfseries,
                  breakable]
\label{p4}
\textbf{You are an intelligent remote sensing analyst.}  
Your task is to generate reasoning-based annotations for Visual Question Answering (VQA) using satellite imagery.

\medskip
\textbf{I will provide:}
\begin{itemize}
    \item A global satellite image (downsampled for efficiency)
    \item The bounding box $[x_{\min}, y_{\min}, x_{\max}, y_{\max}]$ of the \textbf{reference object} mentioned in the question
    \item A natural language question referring to the image
    \item The ground truth answer
\end{itemize}

\medskip
\textbf{Important:}
\begin{itemize}
    \item The reference bounding box corresponds to the \textbf{referent object} in the question (\eg, if the question asks ``What is the structure parallel and closest to the bridge?'', the reference bbox is for the bridge).
    \item The final target answer is derived \textbf{by reasoning relative to this referent object}.
    \item The global image should only be used for context and to explain how one would locate the referent region.
    \item The final answer must be derived \textbf{by analyzing the cropped region corresponding to the referent and its surroundings}.
\end{itemize}

\medskip
\textbf{When writing \texttt{<stage\_1\_reasoning>}, follow these rules:}
\begin{itemize}
    \item At the very start, always begin with:  
    \emph{``This question is asking about <short intent>, therefore I need to crop the image to examine the surroundings of the mentioned target.''}
    \item \textbf{Localization Strategy}: Describe the approximate location of the referent object in natural language (\emph{\eg}, ``the bottom-most long bridge across the river''). Do not output exact coordinates.
    \item \textbf{Reasoning Result}: First output exactly:  
    \emph{``I need to pay attention to the reference object at''}  
    Then output the bounding box in JSON format on the next line:
\begin{lstlisting}[language=json]
[
    {"bbox_2d": [x_min, y_min, x_max, y_max], "label": "<short description of the referent object>"}
]
\end{lstlisting}
    No additional explanation in this section.
\end{itemize}

\medskip
\textbf{Output must strictly follow this structure:}
\begin{lstlisting}[language=json]
<global> - Provide a brief but informative description of the global satellite image (e.g., main structures, spatial layout). </global>

<stage_1_reasoning>
Question Intent: Identify the type of question being asked (e.g., object category, count, color, spatial relation, etc.), and determine what visual information is needed to answer it.

Localization Strategy: Parse the question to identify the referent object (e.g., bridge, river, building cluster). Translate the description into a visual query and locate it in the global image using semantic cues (shape, size, color, spatial arrangement). Summarize the approximate location of the referent in natural language.

Reasoning Result:
I need to pay attention to the reference object at
[JSON bounding box]
</stage_1_reasoning>

<stage_2_reasoning>
Given the cropped region of the referent object, explain how to reason about the final target answer. Specify what visual features or spatial relations should be observed. Clearly connect the reasoning steps from the referent to the final answer.
</stage_2_reasoning>
\end{lstlisting}

\medskip
\textbf{Constraints:}
\begin{itemize}
    \item Do not reveal the final answer in \texttt{<stage\_1\_reasoning>}.
    \item The \texttt{<global>} description must be neutral and avoid giving away the answer.
    \item The \texttt{<stage\_2\_reasoning>} must directly connect the referent to the final target.
\end{itemize}

\medskip
\textbf{Input:}
\begin{lstlisting}[language=json]
Question: {result["question"]}
Ground Truth Answer: {result["ground_truth"]}
Reference Bounding box: {[int(x / scale) for x in hbox]}
\end{lstlisting}

\end{tcolorbox}

\label{sec:rationale}
\subsection{Data Refinement}
Using GPT-4o, we initially generated over 40,000 question–answer pairs. However, several issues were observed during quality inspection. For binary (True/False) questions, the model exhibited a tendency to generate questions whose correct answer was “Yes,” while rarely producing negative cases. For multiple-choice questions, certain answer options were disproportionately favored, for example, “cement” in material-related questions or “summer” in season-related questions.
To address these biases, we manually removed redundant or overly similar questions and supplemented the dataset with additional questions to balance the distribution of answer options. Furthermore, some model-generated outputs contained factual or logical errors, which were corrected through manual revision.
The refinement process was carried out by six annotators, each contributing over ten hours of work. Compared to fully manual annotation of tens of thousands of questions, our proposed semi-automatic data annotation pipeline significantly reduced the annotation workload while maintaining both label accuracy and distributional balance.

\subsection{Dataset Visualization}
\begin{figure*}[h]
    \centering
    \includegraphics[width=0.8\linewidth]{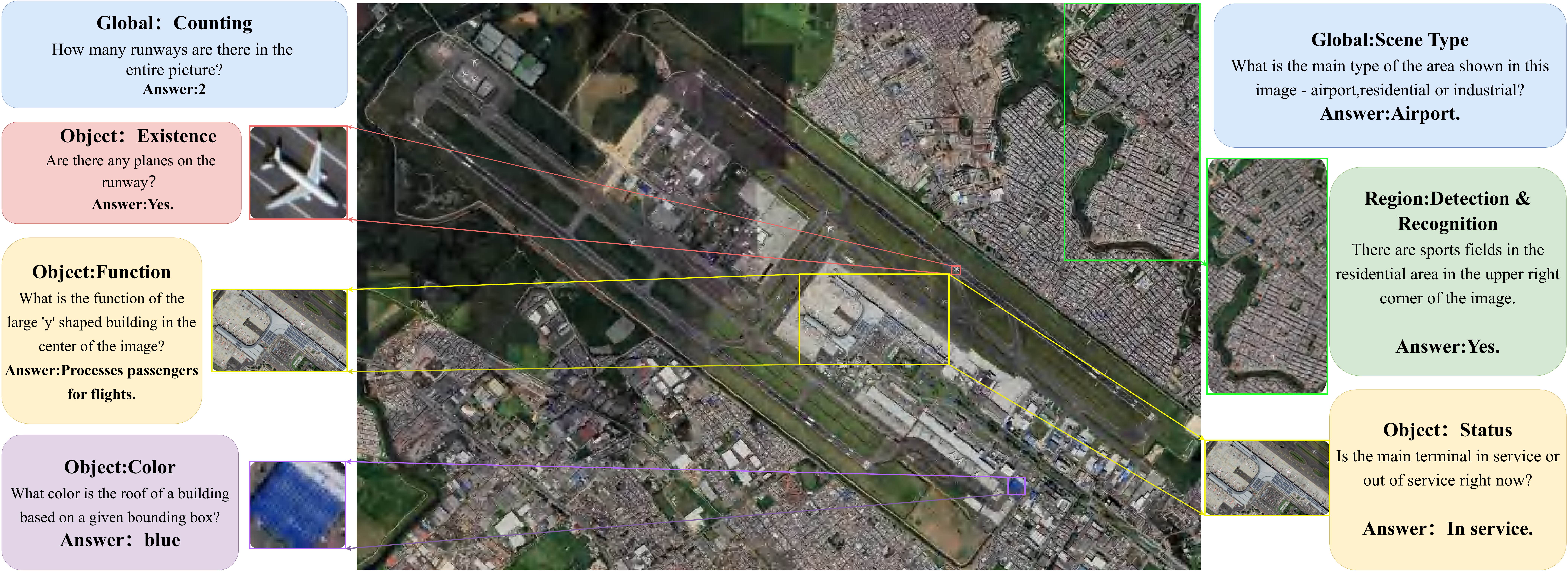}
    \vspace{-2mm}
    \caption{The visualization of examples from LRS-GRO dataset.}
    \label{data_vis1}
    \vspace{-4mm}
\end{figure*}

\begin{figure*}[h]
    \centering
    \includegraphics[width=0.8\linewidth]{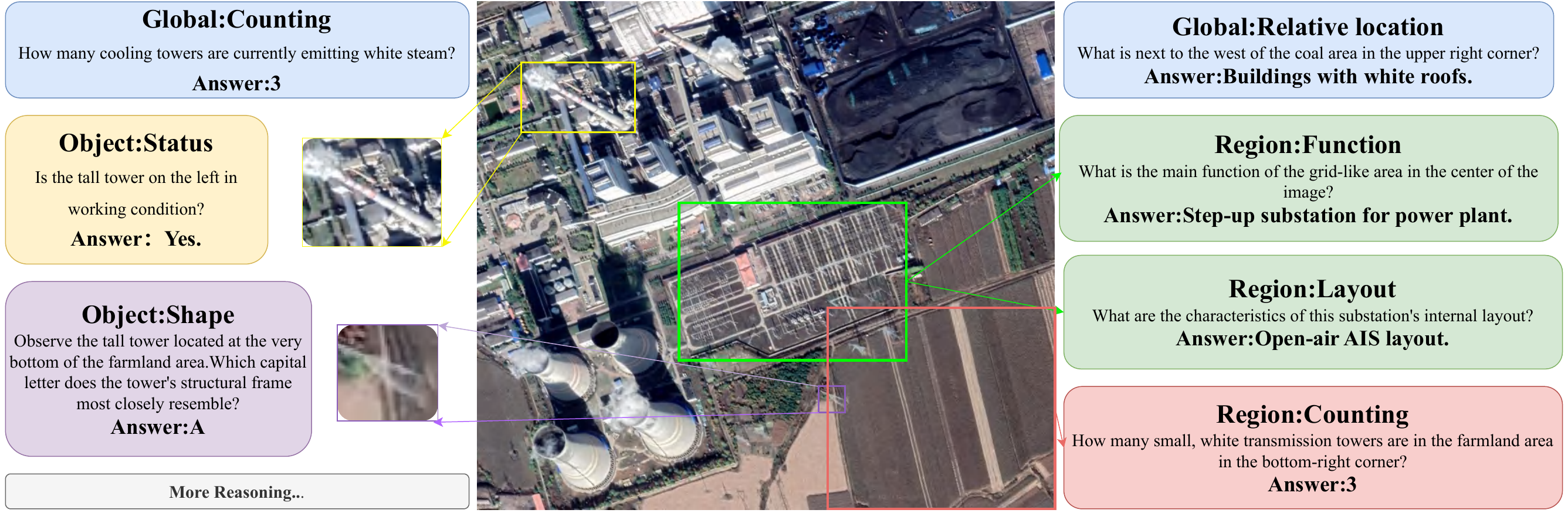}
    \vspace{-2mm}
    \caption{The visualization of examples from LRS-GRO dataset.}
    \label{data_vis2}
    \vspace{-4mm}
\end{figure*}

\begin{figure*}[h]
    \centering
    \includegraphics[width=0.8\linewidth]{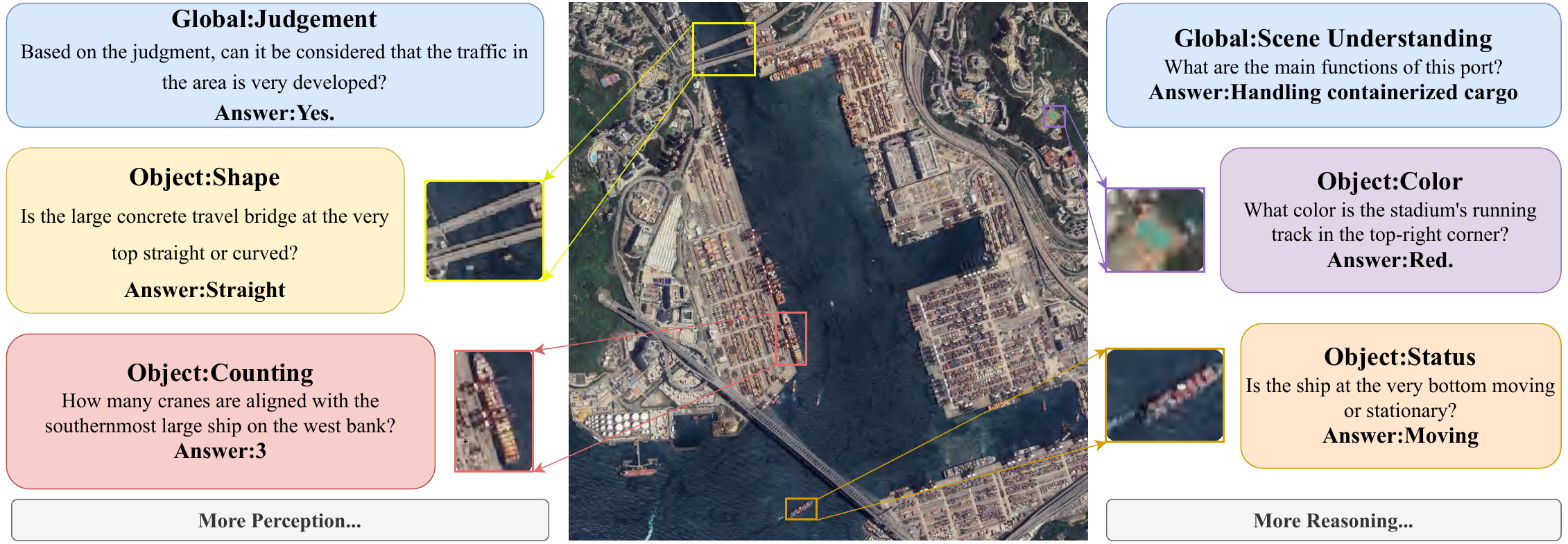}
    \vspace{-2mm}
    \caption{The visualization of examples from LRS-GRO dataset.}
    \label{data_vis3}
    \vspace{-4mm}
\end{figure*}

\begin{figure*}[h]
    \centering
    \includegraphics[width=0.8\linewidth]{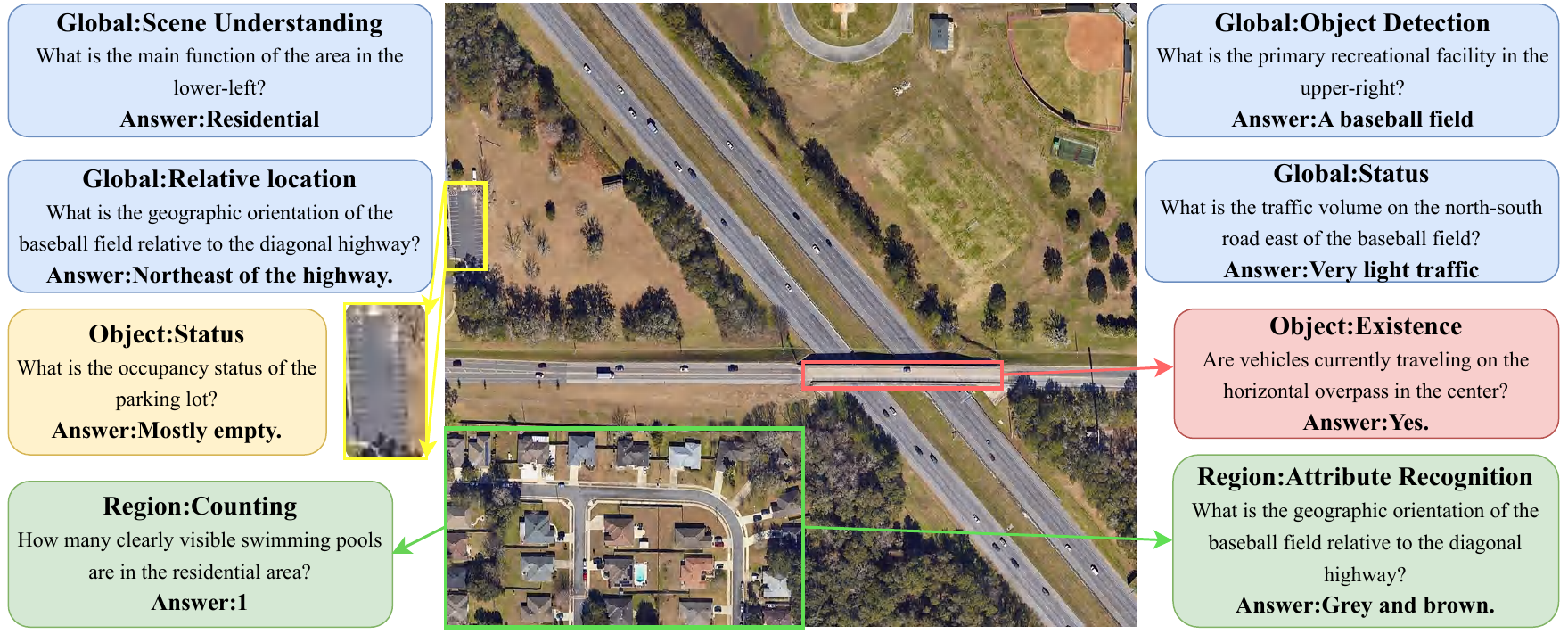}
    \vspace{-2mm}
    \caption{The visualization of examples from LRS-GRO dataset.}
    \label{data_vis4}
    \vspace{-4mm}
\end{figure*}

As shown in Figs~\ref{data_vis1}, \ref{data_vis2}, \ref{data_vis3} and \ref{data_vis4}, we visualize a subset of QA pairs from the LRS-GRO dataset, covering multiple representative geographic landscapes such as airports, factories, ports, bridges, and rural areas. The LRS-GRO dataset provides abundant question–answer pairs and precise bounding box annotations for the global, regional, and object levels.

\subsection{Comparison with LRS-VQA Dataset}
\begin{figure*}[h]
    \centering
    \includegraphics[width=0.7\linewidth]{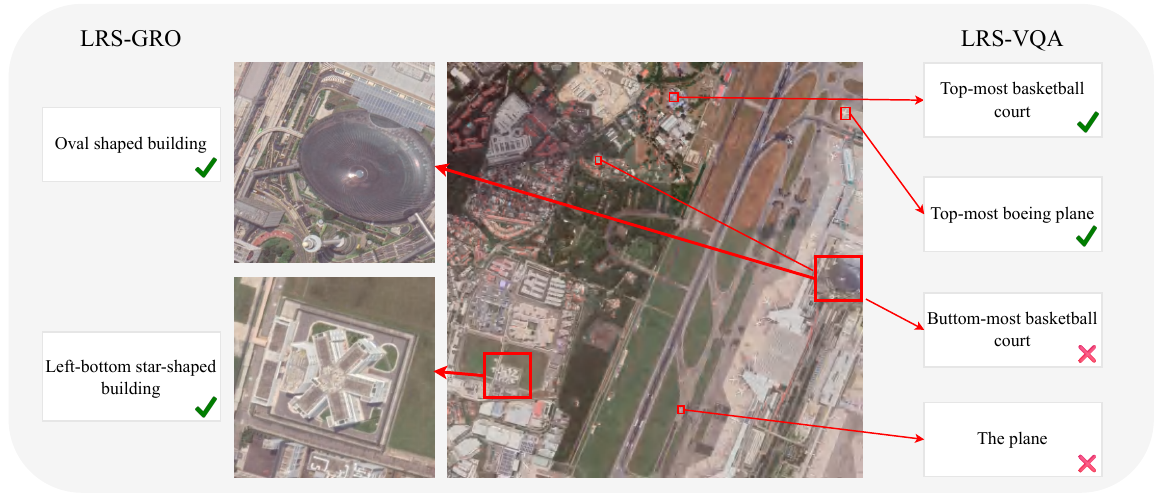}
    \vspace{-2mm}
    \caption{Comparison Between LRS-GRO and LRSVQA.}
    \label{vs1}
    \vspace{-4mm}
\end{figure*}

\begin{figure*}[h]
    \centering
    \includegraphics[width=0.7\linewidth]{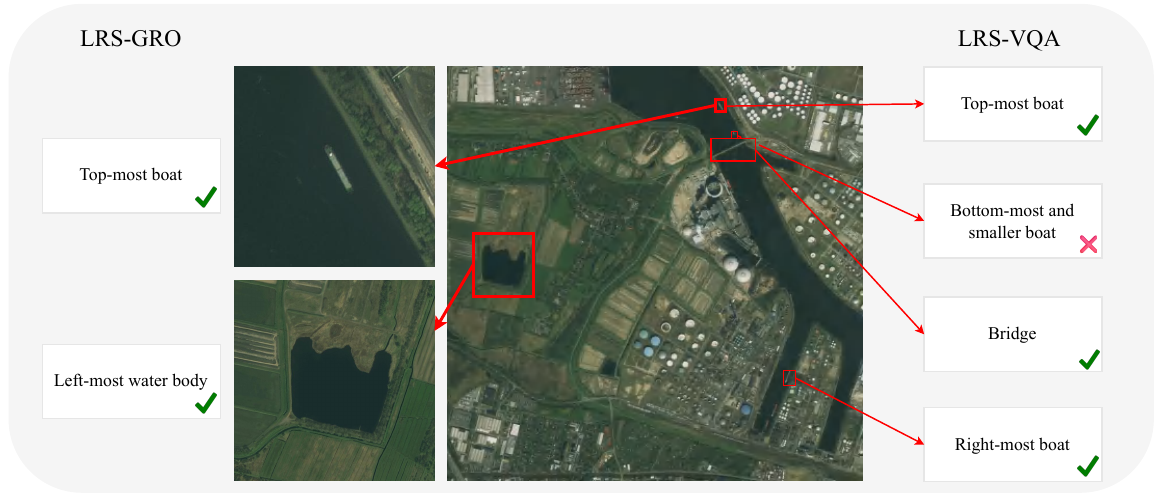}
    \vspace{-2mm}
    \caption{Comparison between LRS-GRO and LRSVQA.}
    \label{vs2}
    \vspace{-4mm}
\end{figure*}

\begin{figure*}[h]
    \centering
    \includegraphics[width=0.7\linewidth]{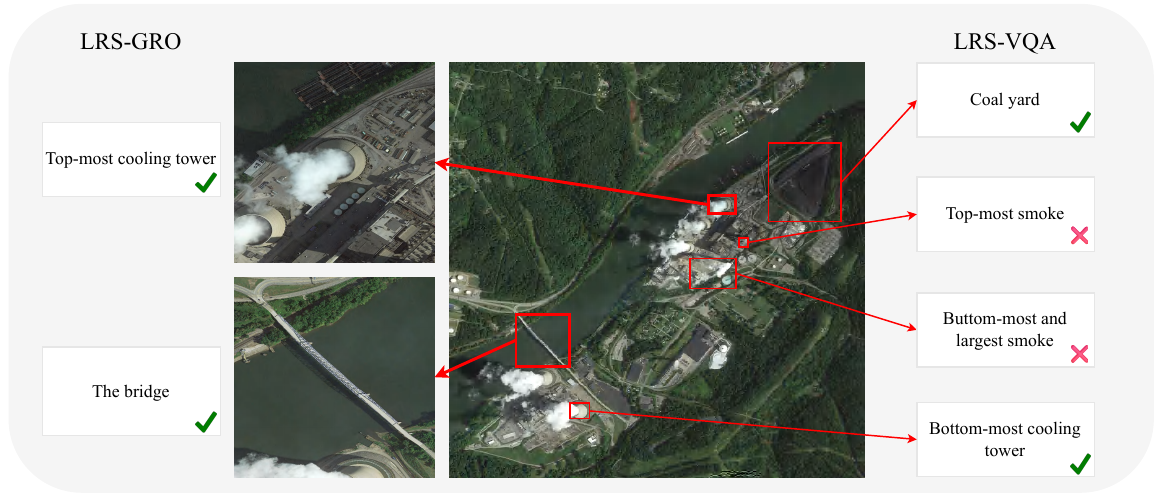}
    \vspace{-2mm}
    \caption{Comparison between LRS-GRO and LRSVQA.}
    \label{vs3}
    \vspace{-4mm}
\end{figure*}

\begin{figure*}[h]
    \centering
    \includegraphics[width=0.7\linewidth]{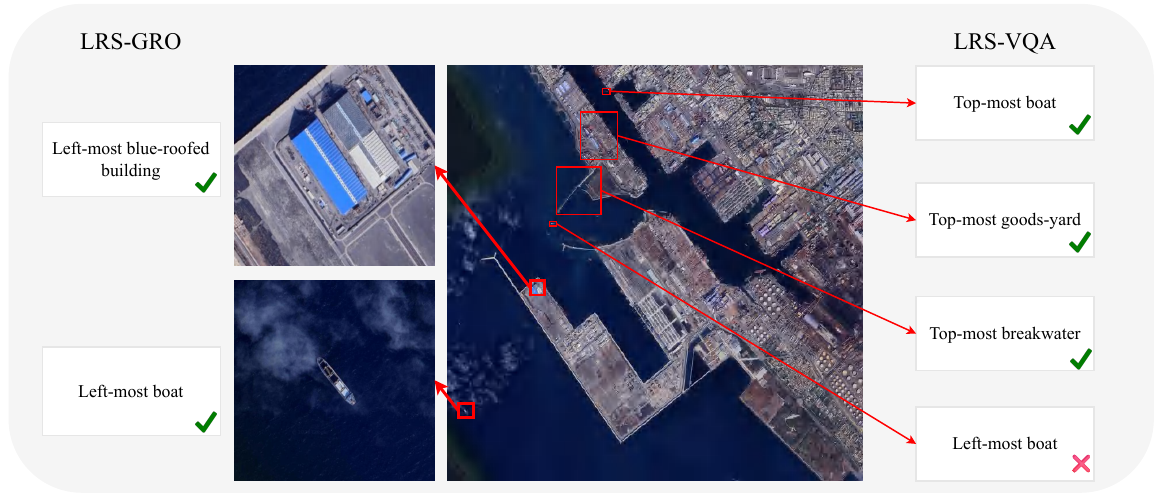}
    \vspace{-2mm}
    \caption{Comparison between LRS-GRO and LRSVQA.}
    \label{vs4}
    \vspace{-4mm}
\end{figure*}

As illustrated in Figs~\ref{vs1}, \ref{vs2}, \ref{vs3} and \ref{vs4}, the LRS-VQA dataset innovatively introduced a GPT-based automated annotation pipeline, offering a novel approach to data labeling. However, due to noticeable hallucinations when GPT processes high-resolution imagery, as illustrated in the figure, incorrect labels may provide the model with misleading rewards during training, thereby hindering convergence. Following the annotation paradigm of LRS-VQA, we further refined and proposed the LRS-GRO dataset. Through meticulous manual annotation and verification, we provide a high-resolution RS image dataset with precise labels.


\section{Training  Details}
\label{appendix:8}

Table~\ref{tab:exp_settings} summarizes the experimental configurations and hyperparameters used for the ZoomEarth under the SFT and GRPO training settings.
\begin{table}[t]
\centering
\scalebox{0.9}{
\begin{tabular}{c cc}
\toprule
\multirow{2}{*}{Configuration} & \multicolumn{2}{c}{VQA}\\
\cmidrule(lr){2-3}
 & SFT & GRPO \\
\midrule
Training component & Full & Full\\
Learning rate & 3e-5 & 1e-7\\
Warmup step & 500 & 50\\
Weight decay & 0.01 & 0.01\\
Batch size & 4 & 32\\
$\alpha$ & - & 200\\
$\beta$ & - & 0.05\\
$\gamma$ & - & 0.04 \\
Optimizer & \multicolumn{2}{c}{AdamW}  \\
Dataset & LRS-GRO/sft 
& LRS-GRO/rl 
\\
Training epoch & 1 & 1\\

\bottomrule
\end{tabular}
}
\caption{Experimental configurations and hyperparameters for SFT and GRPO training. In the GRPO objective, $\alpha$ controls the Region-Guided reward, $\beta$ scales the pattern reward in the overall reward formulation, and $\gamma$ regulates the KL-divergence penalty.}
\label{tab:exp_settings}
\vspace{-6mm}
\end{table}

For a detailed implementation of GRPO, the objective function is defined as follows:
\begin{align}
\mathcal{J}_{\text{GRPO}}(\theta)
&= \mathbb{E}\Big[q\sim P_{\text{sft}}(Q),
\{o_i\}_{i=1}^{G}\sim \pi_{\theta_{\text{old}}}(O|q)\Big] \nonumber\\
&\frac{1}{G}\sum_{i=1}^{G}\frac{1}{|o_i|}
\sum_{t=1}^{|o_i|}
\Bigg[
\hat{A}_{i,t}^{*}-\gamma\mathbb{D}_{\text{KL}}[\pi_{\theta}\,\|\,\pi_{\theta_{\text{ref}}}]
\Bigg]
\end{align}

\noindent
where $P_{\text{sft}}(Q)$ denotes the distribution of queries sampled from the supervised fine-tuning dataset, and $\pi_{\theta}(O|q)$ represents the current policy parameterized by $\theta$, which generates output sequences conditioned on the query $q$. $\hat{A}_{i,j}^{*}$ denotes the clipped advantage reweighted by importance sampling, and $\mathbb{D}_{\text{KL}}[\pi_{\theta}\,\|\,\pi_{\theta_{\text{ref}}}]$ represents an unbiased estimator of the KL divergence. $G$ denotes the number of samples per group, and $|o_i|$ indicates the length of each trajectory. Specifically, $\gamma$ is set to 0.04, and $G$ is set to 4.

\begin{align}
\mathbb{D}_{\text{KL}}[\pi_{\theta}||\pi_{\theta_{\text{ref}}}] =& \frac{\pi_{\theta_{\text{ref}}}(o_{i,t}|q,o_{i,<t})}
     {\pi_{\theta}(o_{i,t}|q,o_{i,<t})} \nonumber\\
 &- \log
\frac{\pi_{\theta_{\text{ref}}}(o_{i,t}|q,o_{i,<t})}
     {\pi_{\theta}(o_{i,t}|q,o_{i,<t})}
 - 1
\end{align}

\begin{align}
    \hat{A}_{i,j}^{*} = \min&\Bigg[\frac{\pi_{\theta}(o_{i,t}|q,o_{i,<t})}
     {\pi_{\theta_{\text{old}}}(o_{i,t}|q,o_{i,<t})}\hat{A}_{i,j}, \nonumber \\
     &\text{clip}\left(\frac{\pi_{\theta}(o_{i,t}|q,o_{i,<t})}
     {\pi_{\theta_{\text{old}}}(o_{i,t}|q,o_{i,<t})}, 1-\epsilon, 1+\epsilon\right)\hat{A}_{i,j}\Bigg]
\end{align}
\begin{equation}
    \hat{A}_{i,j}=\frac{r_i-\text{mean}(r)}{\text{std}(r)}
\end{equation}

The detailed reward components are defined as follows:
\begin{align}
    &r_{IoU} = \begin{cases}
    IOU,&\ \text{if match} \\
    0,& \text{otherwise}
    \end{cases} \\ 
    &r_{R-G} = \text{sigmoid}(\frac{\alpha}{distance+\epsilon}) \\
    &r_{answer} = \begin{cases}
        1,\ & \text{if}\ simularity > 0.8 \\
        simularity, \ &\text{otherwise}
    \end{cases} \\
    &r_{pattern} = \begin{cases}
        1,\ & \text{if match the  patten}\\
        0,\ &\text{otherwise}
    \end{cases}
\end{align}
\begin{equation}
    r=r_{IoU}+r_{R-G}+r_{answer}+\beta r_{pattern}
\end{equation}

Specifically, we set $\epsilon$ to 0.2. The parameter $\beta$ is used to constrain the model outputs to adhere to the predefined format, thereby preventing training collapse. Our experiments show that when $\beta < 0.05$, e.g., $\beta = 0.01$, it fails to take effect and may lead to potential training collapse. Under the premise of stable output formatting, more than $99\%$ of the $r_{\text{pattern}}$ rewards are equal to $1$, and thus increasing $\beta$ has no impact on the overall reward after normalization. Therefore, we set $\beta = 0.05$ in our final configuration.

Finally, the model updates its parameters by performing gradient ascent on the GRPO objective:
\begin{equation}
    \theta \leftarrow \theta + \eta \nabla_{\theta} \mathcal{J}_{\text{GRPO}}(\theta)
\end{equation}
where $\eta$ denotes the learning rate, which is set to $1\times10^{-7}$ in our experiments.

To equip the model with the ability to perform chain-of-thought reasoning in this specific scenario, we applied instruction tuning during both the SFT and RL stages, and used the same instruction prompts during inference to elicit the reasoning capability acquired during training. Specifically, our instruction is as follows:
\begin{tcolorbox}[title=Instruction Used in Training and Evaluation,
                  colframe=black!20,
                  colback=white,
                  coltitle=black,
                  fonttitle=\bfseries,
                  breakable]

\textbf{You are an intelligent remote sensing analyst.}

Given a natural language question about a satellite image, generate a structured reasoning answer as follows:

\begin{enumerate}[label=\arabic*.]
    \item \texttt{<think> ... </think>}
    \begin{itemize}
        \item Provide a neutral one-sentence description of the whole image scene.
        \item Cropping task:\\
        \quad ``This question is asking about \textless short intent\textgreater, therefore I need to crop the image to examine the surroundings of the mentioned target.''
        \item Non-cropping task:\\
        \quad ``This question is asking about \textless short intent\textgreater, therefore I need to analyze the entire image without cropping.''
        \item Include:
        \begin{itemize}
            \item \textbf{Question Intent}: describe the type of question (object category, spatial relation, count, etc.) and the visual information needed.
            \item \textbf{Localization Strategy}:
            \begin{itemize}
                \item Cropping: approximate referent object location in natural language (no coordinates).
                \item Non-cropping: strategy to detect all relevant objects across the entire image.
            \end{itemize}
            \item \textbf{Reasoning Result}:
            \begin{itemize}
                \item \textbf{Cropping}: output exactly one JSON-formatted bbox for the referent, for example:
\begin{lstlisting}[language=json]
[{"bbox_2d": [x_min, y_min, x_max, y_max],
  "label": "<short description>"}]
\end{lstlisting}
                \item \textbf{Non-cropping}: summarize how detected objects will be used to produce the count or answer.
            \end{itemize}
        \end{itemize}
    \end{itemize}

    \item \texttt{<think> ... </think>} (only when the cropped image is provided)
    \begin{itemize}
        \item Explain step-by-step how to reason from the referent (or detected objects) to the final answer.
    \end{itemize}

    \item \texttt{<answer> ... </answer>}
    \begin{itemize}
        \item Provide your final answer as a single word or short phrase.
    \end{itemize}
\end{enumerate}

\medskip
\textbf{Rules:}
\begin{itemize}
    \item Always return exactly one \texttt{<answer>} block.
    \item For tasks that need cropping:
    \begin{itemize}
        \item Provide the bounding box of the object of interest in the first \texttt{<think>} block.
        \item After the cropped image is given, generate another \texttt{<think>} block to derive the answer.
        \item Also include a bounding box in the \texttt{<stage\_2\_reasoning>} block when required.
    \end{itemize}
    \item If unsure about localization, make a best reasonable guess — never state uncertainty.
\end{itemize}

\end{tcolorbox}

\section{Inference Details}
\label{appendix:9}
\subsection{Visualization of the Reasoning Process}
\begin{figure*}
    \centering
    \includegraphics[width=0.82\linewidth]{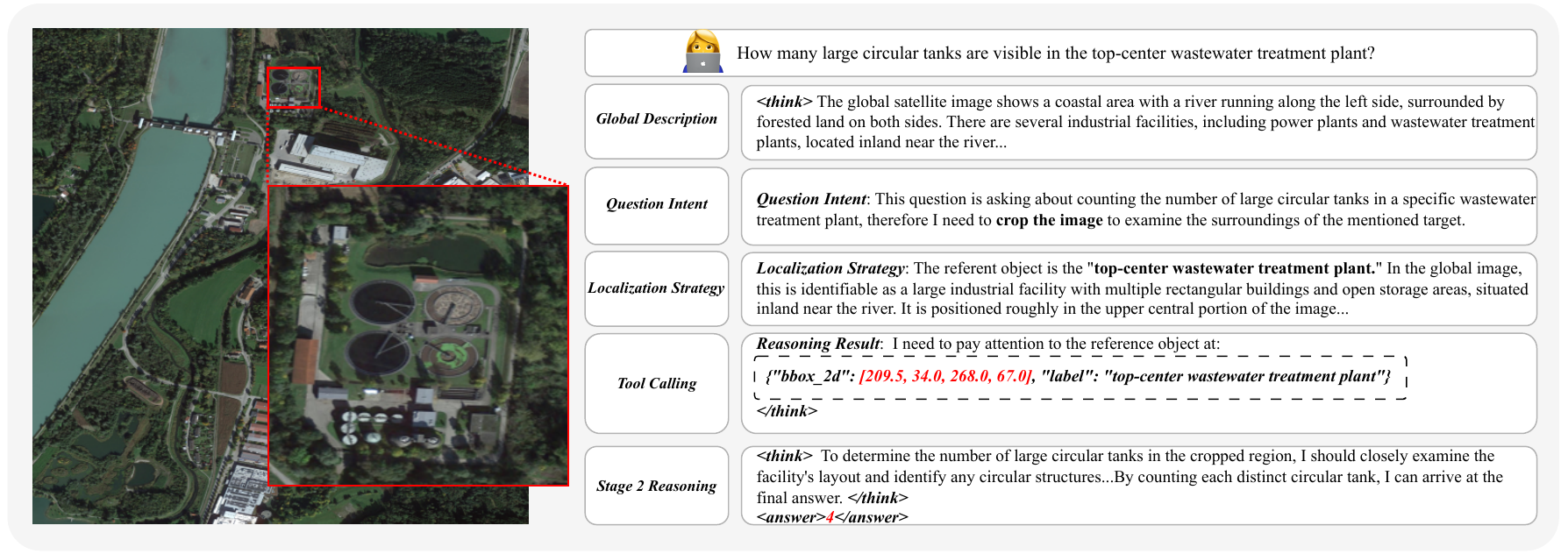}
    \vspace{-2mm}
    \caption{Reasoning process visualization  of ZoomEarth.}
    \label{answer_vis1}
    \vspace{-4mm}
\end{figure*}

\begin{figure*}
    \centering
    \includegraphics[width=0.82\linewidth]{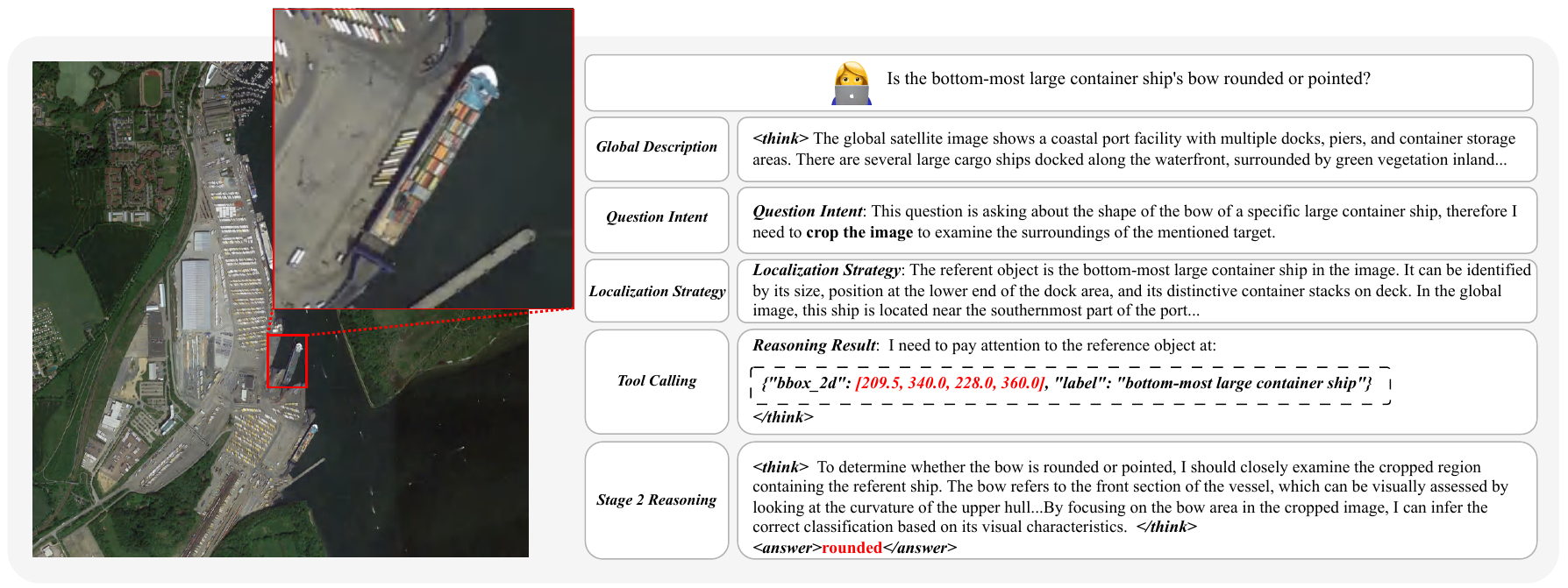}
    \vspace{-2mm}
    \caption{Reasoning process visualization  of ZoomEarth.}
    \label{answer_vis2}
    \vspace{-4mm}
\end{figure*}

\begin{figure*}
    \centering
    \includegraphics[width=0.82\linewidth]{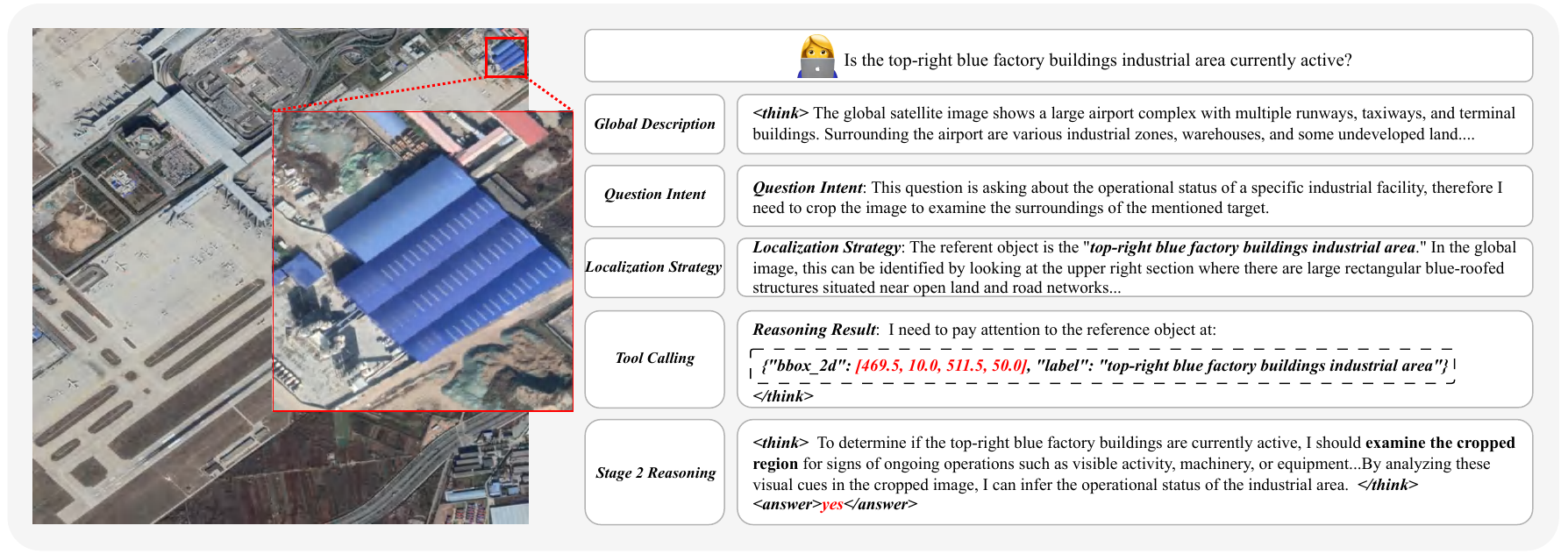}
    \vspace{-2mm}
    \caption{Reasoning process visualization  of ZoomEarth.}
    \label{answer_vis3}
    \vspace{-4mm}
\end{figure*}

\begin{figure*}
    \centering
    \includegraphics[width=0.82\linewidth]{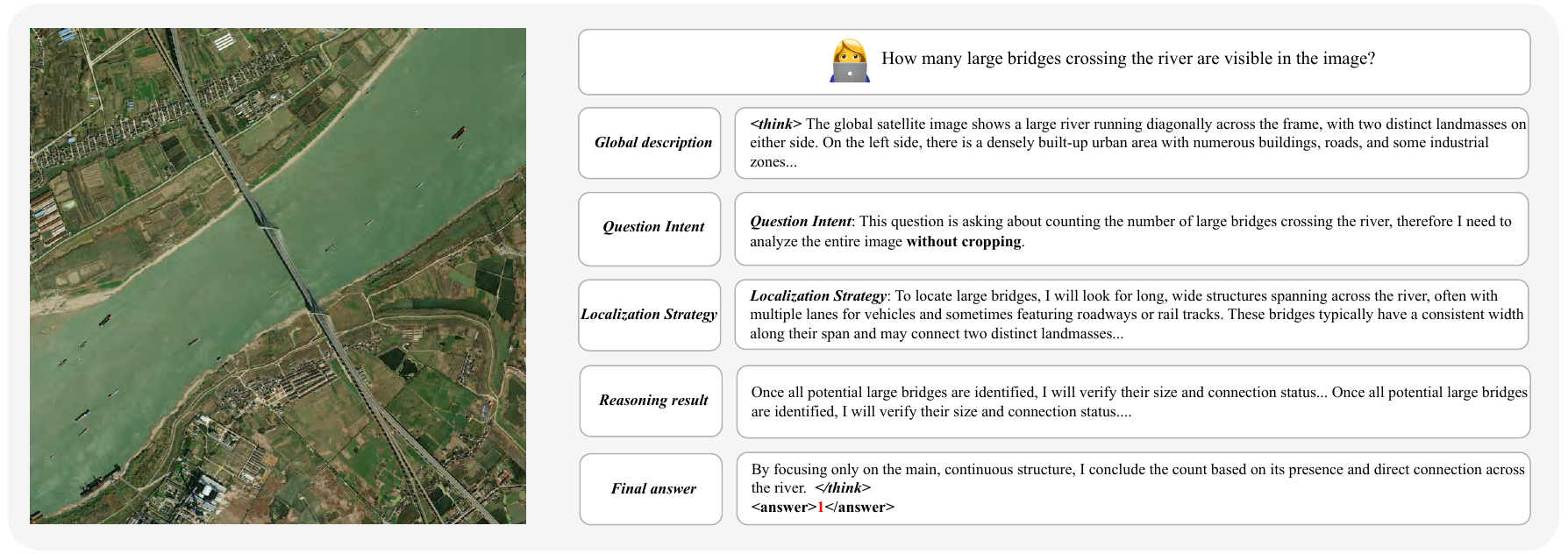}
    \vspace{-2mm}
    \caption{Reasoning process visualization  of ZoomEarth.}
    \label{answer_vis4}
    \vspace{-4mm}
\end{figure*}

As illustrated in the Figs~\ref{answer_vis1}, \ref{answer_vis2}, \ref{answer_vis3} and \ref{answer_vis4}, the reasoning process of ZoomEarth is structured into several sequential stages to ensure both global comprehension and precise localization.

\begin{enumerate}
    \item \textbf{Global Description}: The process begins with a comprehensive global analysis of the input image. This stage aims to establish an overall semantic understanding of the scene, enabling the model to capture contextual relationships and spatial configurations before focusing on specific regions.
    
    \item \textbf{Question Intent Identification}: Following global perception, the model analyzes the user's query to determine its underlying intent. This step isolates the core informational demand, ensuring that subsequent reasoning is aligned with the question's focus.
    
    \item \textbf{ROI Localization Strategy}: Based on the global understanding and the identified question intent, the model performs targeted localization reasoning. This involves determining the spatial regions most relevant to the query and representing them via bounding boxes.
    
    \item \textbf{Tool Calling}: Once ROIs are identified, the model invokes appropriate tools to process the localized areas. This stage is referred to as \textit{secondary perception}, as the model re-examines the image after cropping and zooming the identified regions. The objective is to increase resolution and focus on fine-grained visual details that may be lost in the global view.
    
    \item \textbf{Stage 2 Reasoning}: The refined, high-resolution inputs derived from the localized regions are then used for a second stage of reasoning. This stage integrates global context with localized detail to produce the final answer, ensuring accuracy and relevance to the original query.
\end{enumerate}

This multi-stage process is designed to balance broad contextual awareness with precise visual focus. By first establishing a global understanding and then iteratively narrowing attention to relevant regions, the ZoomEarth model mitigates the risk of missing context while maximizing the accuracy of localized inference.

\subsection{Comparison with Other Models}
\begin{figure}[t]
    \centering
    \includegraphics[width=0.97\linewidth]{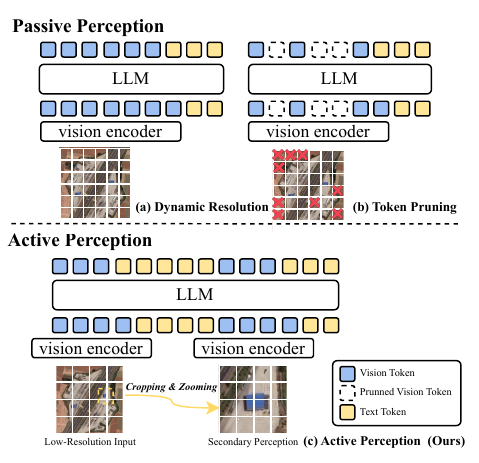}
    \vspace{-4mm}
    \caption{Detailed comparison between passive perception and our proposed active perception method.}
    \label{contrast_appendix}
\vspace{-6mm}
\end{figure}

Many methods have been proposed to address high-resolution image processing, as shown in Fig.~\ref{contrast_appendix}. Techniques such as dynamic resolution and visual token pruning have become mainstream. Dynamic resolution methods first pad the image to an integer multiple of small patches and then split it into a sequence of small patches. This allows the model to handle high-resolution images but does not reduce the number of visual tokens, so a large number of tokens still need to be processed. Visual token pruning removes tokens with low information content or redundant information according to manually defined rules, reducing the number of visual tokens and enabling the model to handle high-resolution images. However, this approach relies on handcrafted rules and has limited generalizability.

Our proposed method first feeds a downsampled low-resolution image into the model. The model then identifies ROIs and calls the cropping-zooming tool to obtain detailed information for these areas. This reduces the total number of visual tokens while preserving fine-grained details in the ROIs.

\begin{figure*}
    \centering
    \includegraphics[width=0.76\linewidth]{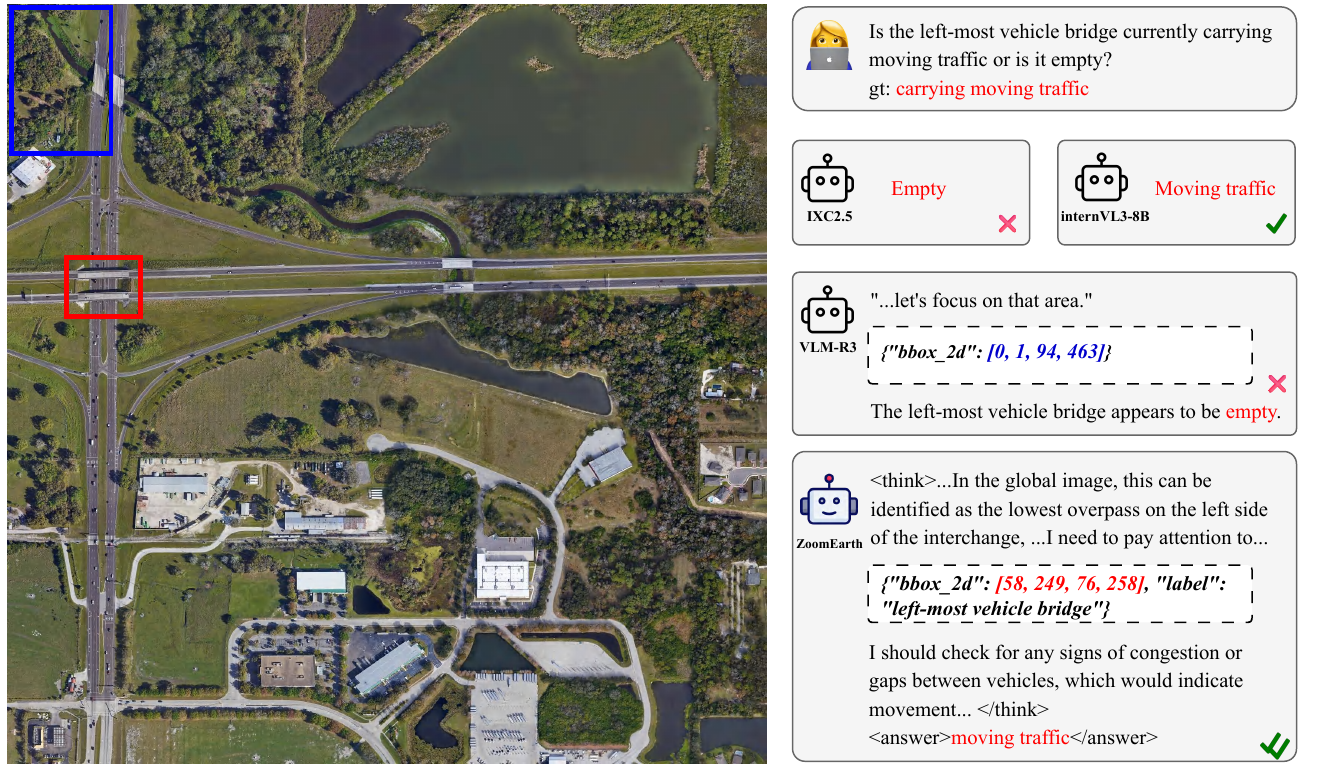}
    \vspace{-2mm}
    \caption{The comparison of the answer between different models.}
    \label{answer_compare_1}
    \vspace{-4mm}
\end{figure*}

\begin{figure*}
    \centering
    \includegraphics[width=0.8\linewidth]{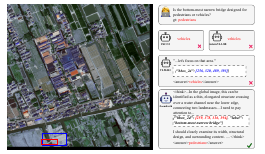}
    \vspace{-2mm}
    \caption{The comparison of the answer between different models.}
    \label{answer_compare_2}
    \vspace{-4mm}
\end{figure*}

\begin{figure*}
    \centering
    \includegraphics[width=0.8\linewidth]{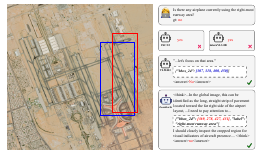}
    \vspace{-2mm}
    \caption{The comparison of the answer between different models.}
    \label{answer_compare_3}
    \vspace{-4mm}
\end{figure*}

As illustrated in the Figs~\ref{answer_compare_1}, \ref{answer_compare_2} and \ref{answer_compare_3}, for questions at the region-level or object-level, general VLMs often produce incorrect answers due to limited visual resolution or hallucinations. Although VLMR\textsuperscript{3} is capable of performing secondary perception, its training methodology is not well-suited for high-resolution RS imagery, rendering it ineffective in this context. In contrast, our proposed ZoomEarth model can accurately localize the region of interest and perform reasoning based on the secondary perception of that region, thereby producing correct answers.

\subsection{Evaluation of APO IoU}

During inference, the model only outputs the region of interest or target, while the answer is often located in the adjacent area outside the ROIs. Therefore, we expand the predicted bounding box to a size of 512 and crop it from the original image. Accordingly, the APO IoU is computed after enlarging both the ground-truth and predicted bounding boxes to a fixed size of 512.

\section{Downstream Tasks}
\subsection{Downstream Instructions}
ZoomEarth can autonomously invoke external tools to perform downstream tasks such as cloud removal, denoising, segmentation, and image editing without additional training. By simply modifying the instruction, the model can be endowed with the ability to utilize these tools effectively. Specifically, the instructions we used in downstream tasks are:

\begin{tcolorbox}[title=Instruction Used for Downstream tasks,
                  colframe=black!20,
                  colback=white,
                  coltitle=black,
                  fonttitle=\bfseries,
                  breakable]

\textbf{You are an intelligent remote sensing analyst.}

Given a natural language question about a satellite image, generate a structured reasoning answer as follows:

\begin{enumerate}[label=\arabic*.]

    \item \texttt{<think> ... </think>}
    \begin{itemize}
        \item Provide a neutral one-sentence description of the whole image scene.
        \item Cropping task: ``This question is asking about \textless short intent\textgreater, therefore I need to crop the image to examine the surroundings of the mentioned target.''
        \item Non-cropping task: ``This question is asking about \textless short intent\textgreater, therefore I need to analyze the entire image without cropping.''
        \item Include:
        \begin{itemize}
            \item \textbf{Question Intent}: describe the type of question (object category, spatial relation, count, etc.) and needed visual information.
            \item \textbf{Localization Strategy}:
            \begin{itemize}
                \item Cropping: approximate referent object location in natural language (no coordinates).
                \item Non-cropping: strategy to detect all relevant objects.
            \end{itemize}
            \item \textbf{Reasoning Result}:
            \begin{itemize}
                \item \textbf{Cropping}: output exactly one JSON-formatted bbox for the referent:
\begin{lstlisting}[language=json]
[{"bbox_2d": [x_min, y_min, x_max, y_max],
  "label": "<short description>"}]
\end{lstlisting}

                \item \textbf{Non-cropping}: summarize how detected objects will be used to produce the count.
            \end{itemize}
        \end{itemize}
    \end{itemize}

    \item \texttt{<tool\_call> ... </tool\_call>}
    \begin{itemize}
        \item After generating the bounding box of the ROI, you must call the tools below.
        \item Follow these formatting rules strictly:
        \begin{itemize}
            \item You must output exactly one \texttt{<tool\_call>} block.
            \item The content inside must be a valid JSON object in the following format:
\begin{lstlisting}[language=json]
<tool_call>
{
    "name": "tool_name",
    "arguments": {
        "arg1": arg1
    }
}
</tool_call>
\end{lstlisting}

            \item Available tools:
            \begin{itemize}
                \item \textbf{Cloud removal:}
\begin{lstlisting}[language=json]
"name": "cloud_removal",
"arguments": {
    "bbox_2d": [x_min, y_min, x_max, y_max]
}
\end{lstlisting}

                \item \textbf{Segmentation:}
\begin{lstlisting}[language=json]
"name": "segmentation",
"arguments": {
    "bbox_2d": [x_min, y_min, x_max, y_max],
    "objects": [obj1, obj2]
}
\end{lstlisting}

                \item \textbf{Denoise:}
\begin{lstlisting}[language=json]
"name": "denoise",
"arguments": {
    "bbox_2d": [x_min, y_min, x_max, y_max]
}
\end{lstlisting}

                \item \textbf{Image editing:}
\begin{lstlisting}[language=json]
"name": "image_editing",
"arguments": {
    "bbox_2d": [x_min, y_min, x_max, y_max],
    "description": "Remove the tree"
}
\end{lstlisting}
            \end{itemize}

            \item Do not include any reasoning or comments outside the \texttt{<tool\_call>} block.
            \item If localization is uncertain, make a best-guess bounding box.
            \item Ensure that the JSON inside \texttt{<tool\_call>} is valid and machine-readable.
        \end{itemize}
    \end{itemize}

    \item Example:
\begin{lstlisting}[language=json]
<tool_call>
{
    "name": "cloud_removal",
    "arguments": {
        "bbox_2d": [120.5, 230.1, 640.3, 880.2]
    }
}
</tool_call>
\end{lstlisting}

\end{enumerate}

\end{tcolorbox}

\subsection{Downstream Examples}
Below are examples of ZoomEarth's response to downstream task's instructions.

\noindent\textbf{Cloud removal}
\begin{tcolorbox}[title=Prompt: Show me the bottom-most blue-roofed building. And then remove the cloud over it using tools. + downstream instruction,
                  colframe=black!20,
                  colback=white,
                  coltitle=black,
                  fonttitle=\bfseries,
                  breakable]
\begin{lstlisting}[language=json]
<think>
The global satellite image shows a large industrial facility surrounded by forested areas and some urban infrastructure. The layout includes multiple rectangular buildings, open storage yards, and adjacent access roads and rail lines. There are also some construction zones and partially visible structures, along with surrounding natural land cover patterns.

Question Intent: This question is asking about identifying the position of a specific blue-roofed building at the bottom-most part of the image, therefore I need to crop the image to examine the surroundings of the mentioned target.

Localization Strategy: The referent object is described as a blue-roofed building located at the bottom-most part of the scene. In the global image, these blue-roofed buildings appear as dark-colored rectangular structures with light-colored roofs, positioned towards the lower end of the industrial complex, near the edge of forested areas and road networks.

Reasoning Result: First, I locate the bottom-most blue-roofed building by identifying the lowest rectangular structure with a lighter roof within the industrial complex. Its position is at the southernmost part of the scene, directly adjacent to forested terrain, and slightly above a cluster of smaller buildings. Then, I crop the image to focus solely on this building to perform further operations.

</think>
    <tool_call>
    {
        "name": "cloud_removal",
        "arguments": {
            "bbox_2d": [379.5, 400.5, 393.0, 411.5]
        }
    }
    </tool_call>
\end{lstlisting}             
\end{tcolorbox}

\noindent\textbf{Segmentation}
\begin{tcolorbox}[title=Prompt: Show me the bottom-most blue-roofed building. And then segment it using tools. + downstream instruction,
                  colframe=black!20,
                  colback=white,
                  coltitle=black,
                  fonttitle=\bfseries,
                  breakable]
\begin{lstlisting}[language=json]
<think>
The global satellite image shows a large industrial facility with multiple buildings, open storage areas, and adjacent forested regions. The layout includes long rectangular structures arranged in a linear fashion, some connected by overpasses or elevated roads, and several smaller buildings scattered around the complex.

Question Intent: This question is asking about identifying a specific blue-roofed building at the bottommost position in the facility, therefore I need to crop the image to examine the surroundings of the mentioned target.

Localization Strategy: The referent object is the bottom-most blue-roofed building. In the global image, this corresponds to a small rectangular structure with a bright blue roof located toward the lower right corner of the facility, near the edge of the wooded area.

Reasoning Result:  
I need to pay attention to the bottom-right section of the global image where a small, distinct blue-roofed structure can be seen. This is positioned at the lower end of the main facility area, above open storage lots, and in close proximity to the forested terrain. The blue roof distinguishes it from other nearby buildings or structures.

</think>
<tool_call>
{
    "name": "segmentation",
    "arguments": {
        "bbox_2d": [340.5, 395.5, 398.0, 461.0],
        "objects": ["bottom-most blue-roofed building"]
    }
} 
</tool_call>
\end{lstlisting}             
\end{tcolorbox}

\noindent\textbf{Denoising}
\begin{tcolorbox}[title=Prompt: Show me the bottom-most blue-roofed building. And then denoise the area around it using tools. + downstream instruction,
                  colframe=black!20,
                  colback=white,
                  coltitle=black,
                  fonttitle=\bfseries,
                  breakable]
\begin{lstlisting}[language=json]
<think>  
The global satellite image shows a large industrial facility with multiple warehouse-like buildings, large storage yards, and elevated conveyor systems. The site is surrounded by hilly terrain with open fields and some scattered clusters of buildings. Road networks and bridges connect different parts of the complex, indicating active transportation links.

Question Intent: This question is asking about identifying the bottom-most blue-roofed building after locating the referent, therefore I need to crop the image to examine the surroundings of the mentioned target.

Localization Strategy: The referent object is described as the bottom-most blue-roofed building. In the global image, blue-roofed structures stand out due to their bright coloration against the lighter-toned roofs and surrounding terrain. The lowest building among them would be the southernmost one in the facility layout, which is distinctly located at the bottom of the image. I can locate it by scanning along the southern edge of the facility and identifying the lowest blue-roofed structure.

Reasoning Result:  
I need to pay attention to the referent object at  
```json
[
    {"bbox_2d": [347.5, 398.5, 398.0, 449.0], "label": "bottom-most blue-roofed building"}
]
```
</think>
<tool_call>
{
    "name": "denoise",
    "arguments": {
        "bbox_2d": [347.5, 398.5, 398.0, 449.0]
    }
}  
</think> 
\end{lstlisting}             
\end{tcolorbox}

\noindent\textbf{Image editing}
\begin{tcolorbox}[title=Prompt: Show me the bottom-most blue-roofed building. And then paint its roof red using tools. + downstream instruction,
                  colframe=black!20,
                  colback=white,
                  coltitle=black,
                  fonttitle=\bfseries,
                  breakable]
\begin{lstlisting}[language=json]
<think> The global satellite image shows a large industrial facility with multiple rectangular buildings arranged in a grid-like pattern across several plots. There are large paved areas, some with clearings, and surrounding the facility are open land and vegetation. Roads connect different sections of the complex within and around the periphery.    

Question Intent: This question is asking about identifying and then visually changing the roof color of a specific blue-roofed building, therefore I need to crop the image to examine the surroundings of the mentioned target.

Localization Strategy: The referent object is described by its roof color - specifically blue. In the global image, blue roofs can be identified among the large rectangular buildings occupying the lower right section of the facility. These buildings have distinctively colored roofs, making them stand out from other structures due to their hue. The blue-roofed building in question is located toward the bottom-right corner of the facility, near the edge where the paved area meets the vegetation.

Reasoning Result:  
I need to pay attention to the reference object at  
```json
[
    {"bbox_2d": [357.5, 409.0, 399.5, 446.0], "label": "bottom-most blue-roofed building"}
]
```
</think>
<tool_call>
{
    "name": "image_editing",
    "arguments": {
        "bbox_2d": [357.5, 409.0, 399.5, 446.0],
        "description": "Paint the roof of the bottom-most blue-roofed building red using appropriate tools."
    }
}  
</think>
\end{lstlisting}             
\end{tcolorbox}






%